%% file: main.tex
\documentclass[]{cit_lab_mfr}

\usepackage{amssymb}
\usepackage{amsmath}
\usepackage{mathtools}
\usepackage{tabularx}
\usepackage{array}
\usepackage{float}
\usepackage{algorithm}
\usepackage{algpseudocode}
\usepackage{pifont}
\usepackage{xurl}

\newcommand{\captionermodel}{OmniFysics-Captioner}
\newcommand{\ppm}{PPM}
\newcommand{\agent}{OmniFysics-Agent}
\newcommand{\bench}{OPC benchmark}
\definecolor{benchmarkgreen}{RGB}{0,150,0}
\definecolor{benchmarkred}{RGB}{220,0,0}
\newcommand{\cmark}{\textcolor{benchmarkgreen}{\ding{51}}}
\newcommand{\xmark}{\textcolor{benchmarkred}{\ding{55}}}

\definecolor{plannerteal}{RGB}{0,78,78}
\definecolor{firstnavy}{RGB}{52,78,138}
\definecolor{followupochre}{RGB}{154,94,28}
\definecolor{audioblue}{RGB}{43,86,145}
\definecolor{visualgreen}{RGB}{45,105,76}
\definecolor{finalburgundy}{RGB}{122,54,72}
\newtcolorbox{plannerprompt}{enhanced standard,breakable,colback=plannerteal!4!white,colframe=plannerteal,colbacktitle=plannerteal,coltitle=white,title={Agent System Prompt},fonttitle=\bfseries,boxrule=0.8pt,arc=2mm,left=5pt,right=5pt,top=4pt,bottom=4pt,before skip=7pt,after skip=7pt}
\newtcolorbox{firstprompt}{enhanced standard,breakable,colback=firstnavy!4!white,colframe=firstnavy,colbacktitle=firstnavy,coltitle=white,title={First-Round Brain Prompt},fonttitle=\bfseries,boxrule=0.8pt,arc=2mm,left=5pt,right=5pt,top=4pt,bottom=4pt,before skip=7pt,after skip=7pt}
\newtcolorbox{followupprompt}{enhanced standard,breakable,colback=followupochre!4!white,colframe=followupochre,colbacktitle=followupochre,coltitle=white,title={Follow-Up Brain Prompt},fonttitle=\bfseries,boxrule=0.8pt,arc=2mm,left=5pt,right=5pt,top=4pt,bottom=4pt,before skip=7pt,after skip=7pt}
\newtcolorbox{audioprompt}{enhanced standard,breakable,colback=audioblue!4!white,colframe=audioblue,colbacktitle=audioblue,coltitle=white,title={Audio Observer Prompt},fonttitle=\bfseries,boxrule=0.8pt,arc=2mm,left=5pt,right=5pt,top=4pt,bottom=4pt,before skip=7pt,after skip=7pt}
\newtcolorbox{visualprompt}{enhanced standard,breakable,colback=visualgreen!4!white,colframe=visualgreen,colbacktitle=visualgreen,coltitle=white,title={Visual Observer Prompt},fonttitle=\bfseries,boxrule=0.8pt,arc=2mm,left=5pt,right=5pt,top=4pt,bottom=4pt,before skip=7pt,after skip=7pt}
\newtcolorbox{finalprompt}{enhanced standard,breakable,colback=finalburgundy!4!white,colframe=finalburgundy,colbacktitle=finalburgundy,coltitle=white,title={Finalizer Prompt},fonttitle=\bfseries,boxrule=0.8pt,arc=2mm,left=5pt,right=5pt,top=4pt,bottom=4pt,before skip=7pt,after skip=7pt}

\title{OmniFysics-Captioner Technical Report: Grounding Omni-Modal Understanding in the Physical World for Better Captioning}
\author[*]{Kaixiang Qiu}
\author[*]{Minghao Han}
\author{Keliang Liu}
\author{Yizhou Liu}
\author{Jinghang Han}
\author{Yue Jiang}
\author{\mbox{Xuecheng Wu}}
\author[\S]{Shunli Wang}
\author[\S]{Lihua Zhang}
\author[\S,\dagger]{Dingkang Yang}

\affiliation{Physical Superintelligence Lab, Fysics AI\\College of Intelligent Robotics and Advanced Manufacturing, Fudan University}

\contribution[*]{Equal contribution}
\contribution[\dagger]{Project lead}
\contribution[\S]{Corresponding Author}

\abstract{Building omni-modal models with physical intelligence requires fine-grained supervision that captures physical evidence such as contact, support, deformation, and state transitions. However, existing omni-modal captioners primarily model general audiovisual semantics and often overlook transient or spatially localized physical evidence. We present a unified framework for physics-aware audiovisual captioning spanning data construction, training, and evaluation. Firstly, we build a data construction pipeline that identifies physics-rich clips and leverages \agent{} to coordinate audio, visual, and physical-perception tools for collecting spatiotemporally aligned and traceable cross-modal evidence; within the Agent, a physical perception model (PPM) fine-tuned on approximately 2M image-level samples serves as a dedicated tool for extracting object-interaction and state-change cues. Secondly, we build the Daily-Physics 50K dataset and introduce the evidence-driven OmniPhysCap (OPC) benchmark to evaluate the recovery of physical and cross-modal evidence from generated captions. Finally, we train \textbf{\captionermodel{}} from the resulting data. Our Captioner matches Gemini 3.1 Pro on audiovisual captioning, achieves state-of-the-art results on multiple video-captioning benchmarks, and substantially outperforms other open-source models. Ablations show that PPM evidence improves physical coverage and produces finer-grained, more reliable cross-modal descriptions.}
\date{August 14, 2026}
\checkdata[Email]{\email{kxqiu26@m.fudan.edu.cn}, \email{dicken@fyscis.ai}, \email{lihuazhang@fudan.edu.cn}}
\checkdata[Project Page]{\url{https://github.com/Fysics-AI/OmniFysics-Captioner}}
\checkdata[Model]{\url{https://huggingface.co/Fysics-AI/OmniFysics-Captioner}}
\checkdata[Dataset \& Benchmark]{\url{https://huggingface.co/datasets/Fysics-AI/OmniPhysics-Caption_benchmark}}

\begin{document}
\makeatletter
\setlength{\@fptop}{0pt}
\setlength{\@dblfptop}{0pt}
\makeatother
\maketitle

\input{paper_content}

\FloatBarrier
\clearpage
\setlength{\bibsep}{2pt plus 0.3ex}

\input{frozen_references}
\clearpage
\appendix

\begingroup
\hypersetup{linkcolor=black}
\setlength{\parindent}{0pt}
\setlength{\parskip}{0pt}
\newcommand{\apptocsection}[2]{%
  \vspace{0.9em}%
  \noindent\textbf{\hyperref[#2]{#1}}\nobreak\dotfill\nobreak\pageref{#2}\par}
\newcommand{\apptocsubsection}[2]{%
  \vspace{0.1em}%
  \noindent\hspace*{1.5em}\hyperref[#2]{#1}\nobreak\dotfill\nobreak\pageref{#2}\par}
\vspace*{1.2em}
{\large\bfseries Appendix Contents\par}
\vspace{1.4em}
\apptocsection{A\quad Video Dataset Details}{app:corpus}
\apptocsubsection{A.1\quad Source Dataset Descriptions}{app:source-datasets}
\apptocsubsection{A.2\quad Event Queries and Clip Construction}{app:event-queries}
\apptocsection{B\quad Agent-Based Caption Generation}{app:agent-details}
\apptocsubsection{B.1\quad Multimodal Evidence Acquisition}{app:evidence-acquisition}
\apptocsubsection{B.2\quad Coverage Completion and Caption Synthesis}{app:coverage-synthesis}
\apptocsection{C\quad Implementation Details}{app:model-details}
\apptocsubsection{C.1\quad Physical Perception Model (PPM)}{app:ppm-details}
\apptocsubsection{C.2\quad Training Configuration}{app:training-configuration}
\apptocsection{D\quad OmniPhysCap (OPC) Benchmark Details}{app:bench}
\apptocsubsection{D.1\quad Independent Evaluation-Video Curation}{app:opc-curation}
\apptocsubsection{D.2\quad Evidence-Grounded Construction}{app:opc-construction}
\apptocsubsection{D.3\quad Caption-to-QA Evaluation Protocol}{app:opc-evaluation}
\apptocsubsection{D.4\quad Human Validation of OPC}{app:opc-human-validation}
\apptocsubsection{D.5\quad Qualitative Case Studies}{app:opc-case-studies}
\apptocsection{E\quad Evaluation Details}{app:evaluation}
\apptocsubsection{E.1\quad Benchmark Overview}{app:benchmark-overview}
\apptocsubsection{E.2\quad Standard Benchmark Evaluation}{app:standard-evaluation}
\apptocsubsection{E.3\quad Controlled Comparisons and Ablations}{app:controlled-comparisons}
\apptocsection{F\quad Limitations and Future Work}{app:limitations-future}
\apptocsection{G\quad The Prompt Design of Agent}{app:prompts}
\endgroup

\clearpage
\input{appendix_content}

\FloatBarrier
\end{document}

%% file: paper_content.tex
\section{Introduction}
\label{sec:intro}

Physical intelligence requires models to move beyond object and scene recognition toward reasoning about object properties, interactions, and state transitions~\citep{yang2025medaide,yang2026toward,qian2026spatialguard,yang2025improving,han2026omnifysics}. This capability underpins embodied decision-making and dynamic world modeling \citep{tung2023physionpp,zheng2024contphy,gao2024physgrounded}. Yet PhysBench~\citep{chow2025physbench} shows that current vision-language models remain weak in physical perception and object-centric reasoning. Although open video datasets contain abundant real-world physical processes, scalable pipelines for discovering and annotating them remain lacking, especially for fine-grained descriptions of properties, interactions, state changes, and outcomes.

\begin{figure*}[!t]
\centering
\includegraphics[pagebox=cropbox,width=1\textwidth]{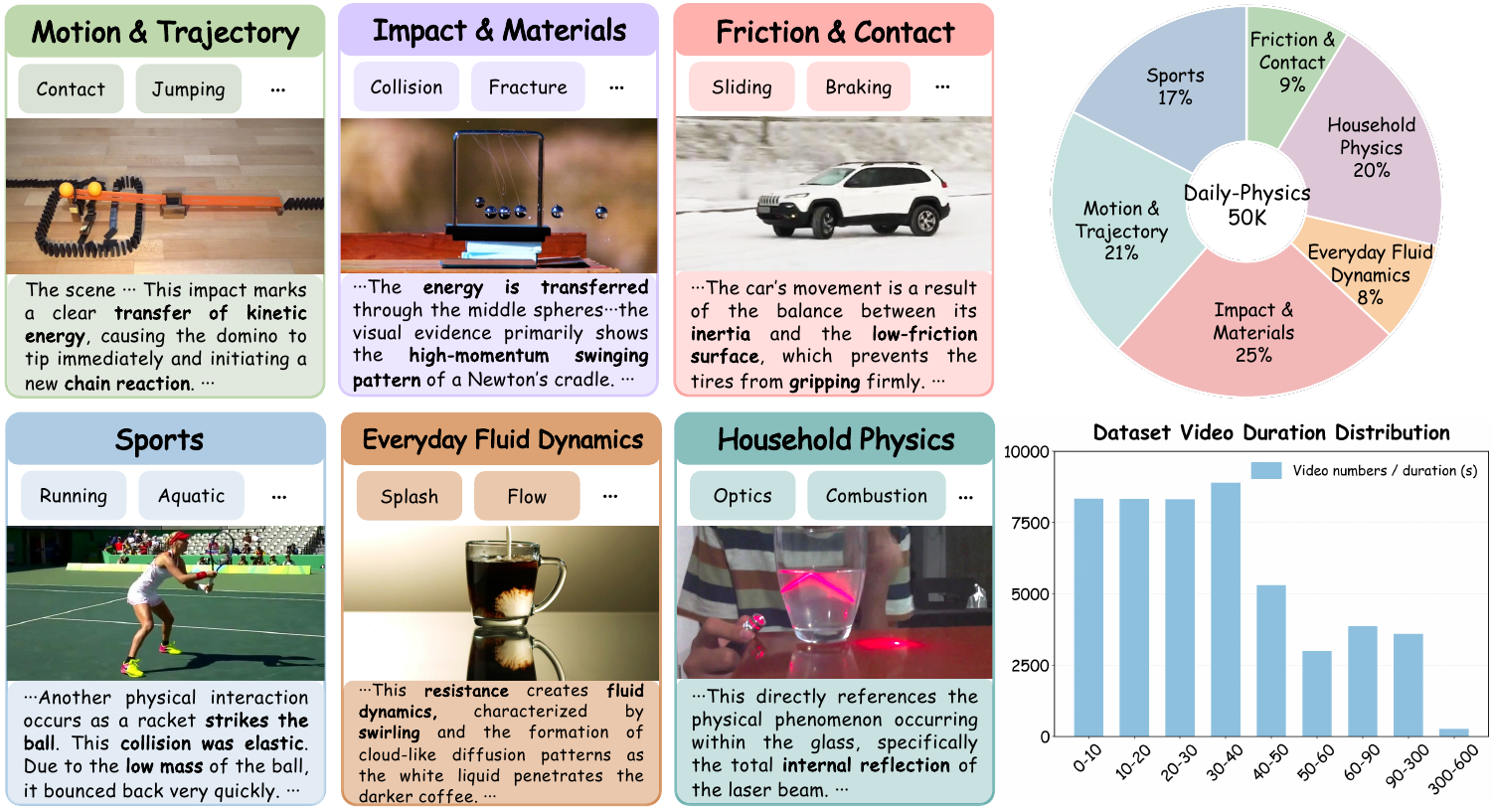}
\caption{Overview of Daily-Physics 50K, representative captions, and data distribution. The dataset spans six categories of everyday physical phenomena; the examples demonstrate our model's physical-perception capabilities, while the charts summarize its physical-event categories and video durations.}
\label{fig:fysics50k-overview}
\end{figure*}

Daily-Physics 50K is a physics-aware audiovisual corpus for detailed video-caption training. It covers six top-level physical-event categories and 23 observable event subcategories, ranging from brief local interactions to extended multi-stage processes. We construct the corpus from heterogeneous video sources through event-query retrieval, clip construction, media-integrity checks, content-quality screening, and deduplication. As shown in Figure~\ref{fig:fysics50k-overview}, the dataset is dominated by clips shorter than one minute while retaining a smaller set of one- to ten-minute processes that capture long-range causal chains, slow state changes, and multi-step interactions.

Video captioning has evolved from retrieval-oriented summaries into a language interface for audiovisual training and reasoning. Omni-modal captioners integrate visual content, speech, music, and environmental sounds, using temporal alignment or agentic data generation to improve cross-modal descriptions \citep{krishna2017dense,chen2024sharegpt4video,chai2025auroracap,tang2025videosalmonn2,chen2025avocado,ma2026omnicaptioner}. Despite strong general-semantic and audiovisual performance, existing methods remain limited along four axes: perceptual granularity for localized cues, spatiotemporal grounding of interactions, physical fidelity in property and causal inference, and cross-modal evidence completeness and traceability. Consequently, they struggle to determine \emph{which object is supported by what}, \emph{how a material responds after contact}, and \emph{what the action changes}. Physics-aware captions must instead bind physical events to specific objects and times while preserving recoverable evidence for downstream reasoning.

Caption construction amplifies limitations: collisions, contact, and deformation are transient, whereas state changes and outcomes span longer intervals and require cross-modal verification. One-pass MLLM annotation can omit local events or hallucinate unobservable properties and causal relations, while tool-augmented methods, including Omni-Detective~\citep{ma2026omnicaptioner} and OmniAgent~\citep{tao2025active}, broaden observation but lack event localization, active evidence acquisition, and traceable aggregation. The evidence-grounding problem requires scalable event selection, targeted cross-modal acquisition, and physically faithful generation.

We develop a data construction and model training framework for omni-modal physical perception. Guided by six physical-event categories, Category-Aware Temporal Anchor Aggregation (CATA) discovers and segments physics-rich clips from heterogeneous video pools. \agent{} uses an active loop to orchestrate multimodal perception tools over localized intervals, yielding cross-modal physical evidence that is spatiotemporally aligned and traceable. We obtain a Physical perception model (PPM) by supervised fine-tuning on approximately 2M image-level physical-perception samples, adding frame-level event and object cues with targeted physical analysis. We introduce Daily-Physics 50K, comprising approximately 50K physics-rich video--caption pairs, and train an end-to-end \captionermodel{} for tool-free captioning. Figure~\ref{fig:fysics50k-overview} illustrates the six-category coverage of Daily-Physics 50K and captions capturing object interactions, material responses, and state changes.

Benchmarks such as VDC~\citep{chai2025auroracap} and Omni-Cloze~\citep{ma2026omnicaptioner} make detailed-caption evaluation objective through QA decomposition or constrained cloze questions, but were not designed for physical perception. They cannot systematically assess whether captions preserve object properties, spatiotemporal interactions, state changes, and physical outcomes. We address this gap with OmniPhysCap (OPC) benchmark, a physics-aware and omission-aware diagnostic benchmark containing 1,000 audiovisual clips and 8,000 questions covering physical interactions and outcomes as well as visual, audio, and audiovisual information. Machine-assisted construction, blind full-video verification, and an explicit \emph{Not Mentioned} option enable \bench{} to distinguish omitted evidence from conflicting evidence in free-form captions. Our contributions are:
\begin{itemize}

    \item We design a pipeline integrating physics-rich clip selection, active \agent{} evidence acquisition, and caption generation. Daily-Physics 50K comprises approximately 50K video-caption pairs and will be released.
    \item We introduce OmniPhysCap (OPC) benchmark to standardize the evaluation of physical and cross-modal information retained by generated captions. The benchmark has been publicly released.
    \item Using Daily-Physics 50K, we train \captionermodel{}, which achieves state-of-the-art performance on VDC Detailed, Omni-Cloze, and caption-to-QA cascade evaluations across multiple omni-modal benchmarks. Our model has also been publicly released.
    \item Through experiments and ablations, we demonstrate that explicit physical perception strengthens fine-grained omni-modal understanding by improving the coverage of object properties, physical interactions, and state changes.
\end{itemize}

\section{Related Work}

\noindent\textbf{Omni-modal Video Reasoning.}\quad
Video MLLMs have progressed from visual temporal modeling toward omni-modal understanding that jointly processes visual content, speech, and environmental sounds. Models such as VideoLLaMA~2 \citep{cheng2024videollama2}, Qwen2.5-Omni \citep{xu2025qwen25omni}, and Qwen3-Omni \citep{qwen2025omni} continue to improve joint audiovisual modeling. For fine-grained captioning, Tarsier2 \citep{yuan2025tarsier2}, AuroraCap \citep{chai2025auroracap}, video-SALMONN~2 \citep{tang2025videosalmonn2}, UGC-VideoCaptioner \citep{wu2025ugc}, and AVoCaDO \citep{chen2025avocado} advance video captioning through data construction, preference optimization, and temporal fusion; Omni-Captioner \citep{ma2026omnicaptioner} uses multi-round tool calls to generate detailed supervision, while the recent AVSCap \citep{wang2026avscap} and TCA-Captioner \citep{zhao2026tca} further emphasize audiovisual event binding and temporal alignment. Correspondingly, evaluation has shifted from surface-form similarity toward QA decomposition in VDC \citep{chai2025auroracap}, temporal reasoning in Daily-Omni \citep{zhou2025dailyomni}, and information recovery in Omni-Cloze \citep{ma2026omnicaptioner}. However, existing methods primarily emphasize general details and modality coverage, often reducing collisions, deformation, and state changes to action labels without explaining the object interactions underlying auditory and visual changes. Physical perception is therefore essential for moving omni-modal models beyond information aggregation toward understanding real-world dynamic processes.

\noindent\textbf{Physical Perception and Evaluation.}\quad
Physical understanding requires models to recognize object properties, interaction relations, and continuous state changes. Physion++ \citep{tung2023physionpp} and ContPhy \citep{zheng2024contphy} study the inference of latent physical properties from dynamic interactions, while Physically Grounded VLM \citep{gao2024physgrounded} and PACS \citep{yu2022pacs} extend physical concepts to embodied manipulation and audiovisual commonsense. PhysBench \citep{chow2025physbench} systematically evaluates object properties, relations, and dynamics; MVP \citep{krojer2025mvp} reduces reasoning shortcuts through minimal video pairs; and MASS \citep{wu2025mass} emphasizes spatiotemporal grounding in physical reasoning. More recent PhysGame \citep{cao2024physgame} and PhysicsMind \citep{mak2026physicsmind} further extend evaluation to physical anomaly recognition and mechanics reasoning in real-world and simulated scenes. However, these benchmarks largely rely on predefined question answering or outcome prediction, whereas existing captioning benchmarks focus on general details. A dedicated evaluation of physical-evidence recoverability in open-ended omni-modal descriptions is still lacking. This work aims to fill this gap.

\begin{figure*}[!t]
\centering
\includegraphics[width=1\textwidth]{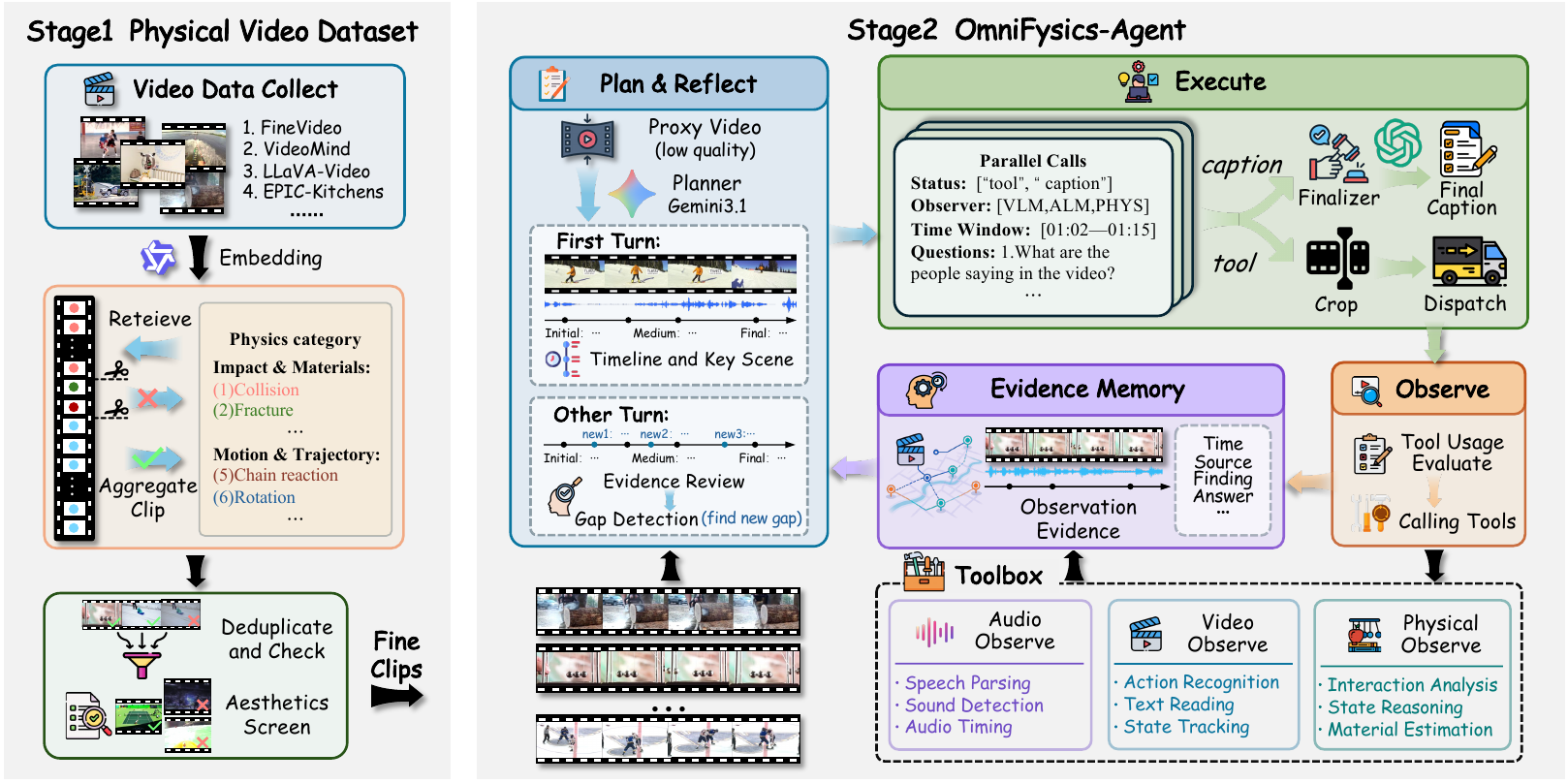}
\caption{Pipeline of our caption work. Stage 1 selects and filters physics-rich clips from heterogeneous video collections. Stage 2 uses \agent{} to coordinate multimodal tools, aggregate traceable evidence, and generate detailed captions.}
\label{fig:overview}
\end{figure*}

\section{Methodology}
\label{sec:method}

As shown in Figure~\ref{fig:overview}, we first select approximately 50K video clips covering diverse physical events. \agent{} then aggregates audio, visual, and \ppm{} evidence through multi-round tool calls and generates detailed captions; together, these clips and captions form Daily-Physics 50K. We further fine-tune Qwen3-Omni on Daily-Physics 50K to obtain an end-to-end omni-modal Captioner that directly processes raw audiovisual input. In addition, we curate a clip-level held-out set of 1K videos, removing exact and perceptual near-duplicates of the training clips, to construct OmniPhysCap (OPC) benchmark, which evaluates how well captions retain physical and cross-modal information.

\subsection{Physical Video Dataset}
\label{sec:method-data}

\noindent\textbf{Event Definitions and Candidate Discovery.}\quad
To build a video pool for physical perception and understanding, we select nine datasets: AVQA~\citep{li2022musicavqa}, Bilibili-Videos, EPIC-Kitchens~\citep{damen2022epickitchens100}, LLaVA-Video~\citep{zhang2025llavavideo}, LU-AVS~\citep{liu2024luavs}, VideoMind~\citep{yang2025videomind}, WISA~\citep{wang2025wisa}, FineVideo~\citep{huggingface2024finevideo}, and unAV-100~\citep{geng2023unav100}. Together, these sources provide broad coverage across content domains, viewpoints, and temporal scales, including audiovisual correspondence, human--object interaction, physical phenomena, and multi-event processes in both short and long videos. Following an organization based on physical phenomena observable in everyday life, we define six top-level categories and 23 event subcategories spanning sports biomechanics, motion and trajectories, impact and material response, everyday fluid dynamics, household thermodynamics and optics, and surface friction and contact. Their event definitions are used as candidate-retrieval queries; the complete taxonomy is provided in the Appendix.

\noindent\textbf{Physical-Clip Retrieval and Segmentation.}\quad
Invoking a generative VLM on every video to identify events and predict boundaries would be costly. We instead propose \emph{Category-Aware Temporal Anchor Aggregation} (CATA). First, Qwen3-VL-Embedding-8B \citep{li2026qwen3vlembedding} encodes event queries and reusable visual features from frames sampled at 1 fps. Frame--query similarities then produce high-scoring temporal anchors with event labels. Finally, CATA merges adjacent anchors from the same category and starts a new segment when the category changes or the inter-anchor gap exceeds a threshold. Candidate clips are 3--60 seconds long, while a subset of 1--10-minute long-process videos is selected. After stratified sampling, deduplication, media-integrity checking, content-quality inspection, and visual-aesthetic evaluation, approximately 50K clips are used for Agent annotation and training. A human evaluation of randomly sampled clips shows that 98.2\% are temporally complete and 96.9\% contain physical events consistent with their assigned categories. A separately curated, clip-level held-out set of 1K clips constitutes the \bench{} test set. The Appendix reports the parameters and data distribution.

\subsection{\agent{}}
\label{sec:method-agent}

To generate physics-rich supervision from fine-grained audiovisual evidence, we introduce the active-perception agent shown in Figure~\ref{fig:overview}. Unlike a preset tool chain or indiscriminate full-media invocation \citep{yao2023react,tao2025active,ma2026omnicaptioner}, \agent{} first forms a global event timeline from a low-cost audiovisual proxy, locates local intervals that require further inspection, and dynamically orchestrates modality-specific tools. Each observation batch is written to Evidence Memory and drives the next Plan--Execute--Observe--Reflect round, progressively refining modality choice, temporal scope, and question focus.

\noindent\textbf{Modality-Specific Observer Toolbox.}\quad
The toolbox exposes three complementary interfaces. An audio large-language model (ALM) tool parses speech, music, and environmental sounds in local clips and extracts their temporal order; a vision large-language model (VLM) tool verifies local visual events, OCR content, and state changes before and after each event; and \ppm{} focuses on objects and physical phenomena in representative frames, perceiving physical cues such as material, contact, and deformation, and analyzing object interactions, state changes, and their potential outcomes. Together, the three tools help the Agent supplement and cross-validate critical evidence, enabling more accurate and detailed video descriptions consistent with physical laws.

We represent each tool call as $a=(m,\tau,\mathbf q)$, where $m\in\mathcal T=\{\mathrm{ALM},\mathrm{VLM},\ppm\}$ selects an Observer, $\tau$ is a short interval localized in the original video, and $\mathbf q$ is a list of modality-directed questions about that interval. This tuple is the basic unit used by the Planner to form batches and by the Executor to route local media. All outputs are stored as source-attributed candidate evidence.

To enhance the Agent's ability to perceive, recognize, and describe physical information in visual scenes, we perform supervised fine-tuning on approximately 2M image-level physical-instruction examples to obtain \ppm{}. The model can identify the physical properties and states of objects, understand spatial relationships and interaction cues among objects, and convert observable motion trends, state changes, and their potential outcomes into fine-grained textual descriptions. It thereby provides a foundation for the Caption Agent to generate descriptions with richer physical information and stronger factual support. Data construction and training details are provided in the Appendix.

\noindent\textbf{Evidence-Guided Batch Planning.}\quad
Given a video $V$, the system constructs a low-cost audiovisual proxy $\widetilde V$ for global navigation. Let $M_r$ be Evidence Memory and $b_r$ the residual call budget at the beginning of round $r$. The Planner proposes only the candidate batch for the current round:
\begin{equation}
\widehat Q_r=\pi_\psi(\widetilde V,M_r,b_r).
\label{eq:agent-plan}
\end{equation}
Here $\pi_\psi$ is the structured Planner and $\widehat Q_r$ contains multiple calls $a$. The first round builds a cross-modal event skeleton from the proxy and selects short intervals with high expected information value. Later rounds condition on accumulated evidence to refine modality, interval scale, and question focus, or to stop. Before execution, deterministic $\operatorname{Admit}$ validates requests and forms the executable batch $Q_r$ under Observer availability and the residual budget.

\noindent\textbf{Batched Execution and Evidence Update.}\quad
For each admitted call, the Executor crops the corresponding short interval from the original audiovisual stream and routes the local media and directed questions to the selected Observer. Calls in the same round execute in parallel. This hierarchy uses the proxy for global navigation and original local media for fine-grained verification, concentrating observation compute on information-dense moments while adapting the tool mix and temporal granularity across rounds:
\begin{equation}
O_r=\operatorname{Execute}(Q_r;V).
\label{eq:agent-observe}
\end{equation}
$O_r$ is the observation batch corresponding to $Q_r$. Once the full batch returns, Evidence Memory advances as $M_{r+1}=\operatorname{Update}(M_r,Q_r,O_r)$.

\noindent\textbf{Reflection, Backfill, and Evidence Synthesis.}\quad
The updated $M_{r+1}$ becomes the next Planner input, allowing the same model to reassess modality complementarity, temporal localization, and question focus before issuing follow-up calls or proposing termination. This closes the active-perception loop. After the active loop exits, Coverage no longer invokes the Planner. It executes deterministic default observations according to video duration, modality availability, and prior calls within the residual budget. The Finalizer then accesses no media and issues no tool calls; it organizes the final Evidence Memory and configuration-selected non-media planning context $\mathcal P^\star$ into a temporally coherent and nonredundant caption. Algorithm~\ref{alg:agent-generation} summarizes the complete control flow.

\begin{figure}[!t]
\centering
\begin{minipage}{0.65\textwidth}
\hrule
\vspace{2pt}
\refstepcounter{algorithm}
\label{alg:agent-generation}
\noindent\textbf{Algorithm~\thealgorithm} \agent{} Inference Process
\vspace{2pt}
\hrule
\vspace{3pt}
\small
\begin{algorithmic}[1]
\Require video $V$; Observers $\mathcal T$; call/round budgets $B,R_{\max}$
\Ensure caption $C_A$, final Evidence Memory $M^\star$
\State $\widetilde V\gets\operatorname{Proxy}(V)$; $M\gets\varnothing$; $b\gets B$
\For{$r=1,\ldots,R_{\max}$}
    \If{$b=0$}
        \State \textbf{break}
    \EndIf
    \State $\widehat Q_r\gets\pi_\psi(\widetilde V,M,b)$ \Comment{initial plan or evidence reflection}
    \State $Q_r\gets\operatorname{Admit}(\widehat Q_r;M,\mathcal T,b)$
    \If{$Q_r=\varnothing$}
        \State \textbf{break}
    \EndIf
    \State $O_r\gets\operatorname{Execute}(Q_r;V)$ \Comment{parallel within the batch}
    \State $M\gets\operatorname{Update}(M,Q_r,O_r)$
    \State $b\gets b-|Q_r|$
\EndFor
\State $M^\star\gets\operatorname{Coverage}(V,M,b;\mathcal T)$ \Comment{deterministic; no Planner}
\State $C_A\gets\operatorname{Finalize}(M^\star,\mathcal P^\star)$ \Comment{no media or tools}
\State \Return $(C_A,M^\star)$
\end{algorithmic}
\vspace{2pt}
\hrule
\end{minipage}
\end{figure}

\subsection{\captionermodel{}}
\label{sec:method-captioner}

The active acquisition and synthesis pipeline yields 50K evidence-rich, physics-dense video--caption pairs. To amortize the agent's observation and evidence-organization capability into a single forward pass, we fully fine-tune Qwen3-Omni-30B-A3B-Instruct on raw audiovisual input with agent captions as supervision \citep{qwen2025omni}. Let $\mathcal D_{\rm cap}=\{(V_i,C_{A,i})\}_{i=1}^{N}$ denote this dataset, where $N=50\mathrm K$.

The model jointly reads visual stream $X^v$ and audio stream $X^a$ from the video container and minimizes the conditional negative log-likelihood:
\begin{equation}
\mathcal L_{\rm cap}(\theta)
=-\frac{1}{N}\sum_{i=1}^{N}
\log p_\theta\!\left(C_{A,i}\mid X_i^v,X_i^a\right).
\label{eq:caption-loss}
\end{equation}
Equation~\eqref{eq:caption-loss} uses teacher forcing, with loss computed only on assistant-caption tokens. Supervision contains only the final caption, not temporal windows, tool requests, Observer returns, or reflection trajectories. The Captioner amortizes evidence organization and description from the Agent's final output, rather than learning an explicit planning or tool-use policy. At deployment, it generates a caption from raw audio and video in one forward pass without external tools.

\subsection{OmniPhysCap (OPC) Benchmark}
\label{sec:method-bench}

\noindent\textbf{Benchmark Overview.}\quad
Existing detailed-caption benchmarks provide strong visual-reference, event-level, or cloze-based evaluation, but still lack systematic evaluation of omni-modal information, especially physical interactions and outcomes. As summarized in Table~\ref{tab:caption-benchmark-comparison}, \bench{} complements these benchmarks with typed diagnostics for physics and audiovisual omission. It contains 1,000 audiovisual clips of 6--60 seconds and 8,000 multiple-choice probes, with 5--10 probes assigned per video to provide more precise and appropriate coverage than a fixed question count. Together, these probes cover eight information types, including general semantics, temporal relations, audio, audiovisual alignment, and physical interactions and outcomes.

\setlength{\textfloatsep}{14pt plus 2pt minus 2pt}
\begin{table}[!t]
\centering
\small
\setlength{\tabcolsep}{3.2pt}
\captionof{table}{Comparison with detailed video-caption evaluation benchmarks. ``Units'' denotes semantic items used to inspect a generated caption, and ``Omission'' denotes an explicit absent-information option. V and A denote visual and audio input; MCQ denotes multiple-choice question.}
\label{tab:caption-benchmark-comparison}
\resizebox{\textwidth}{!}{%
\begin{tabular}{@{}lccccccc@{}}
\toprule
\textbf{Benchmark} & \textbf{Videos} & \textbf{Units} & \textbf{Input} & \textbf{Physics} & \textbf{Audio/AV} & \textbf{Omission} & \textbf{Evaluation design} \\
\midrule
VDC Detailed~\citep{chai2025auroracap} & 1,027 & 19,811 & V & \xmark & \xmark & \xmark & Open QA over five structured reference views \\
DREAM-1K~\citep{wang2024tarsier} & 1,000 & 6,298 & V & \xmark & \xmark & \xmark & Bidirectional entailment over atomic events \\
Omni-Cloze~\citep{ma2026omnicaptioner} & 2,320 & 69,600 & V+A & \xmark & \cmark & \cmark & Fixed-count caption-to-cloze; 30 blanks/video \\
\textbf{\bench{}} & 1,000 & 8,000 & V+A & \cmark & \cmark & \cmark & Adaptive typed MCQ; video-macro score \\
\bottomrule
\end{tabular}%
}
\vspace{-8pt}
\end{table}

\noindent\textbf{Evidence-Grounded Question Construction.}\quad
To improve questions, answers, and distractors reliability, we use Gemini 3.1 Pro to perform multi-round consistency checks and replace invalid items using the complete audiovisual input. Candidates that are ambiguous, insufficiently supported, or internally inconsistent are filtered based on answer consistency, option mutual exclusivity, and support from video evidence, reducing hallucinations introduced by automatic question generation. Verified questions are then semantically deduplicated, constrained for content coverage, and cross-validated by human annotators to form the final question set. Unlike cloze benchmarks using lexical blanks, \bench{} retains semantically distinct, event-grounded questions, prioritizing diagnostic coverage over volume.

\noindent\textbf{Caption-to-QA Evaluation Protocol.}\quad
To ensure consistent cross-system comparison, each system generates one detailed caption for every test video under the same media input. Reported scores use video-level macro-averaging across test clips. Formal evaluation uses GPT-5.6 as the fixed caption-only Judge. To explicitly characterize information missing from captions and reduce guessing bias introduced by closed-set selection, we add a \emph{Not Mentioned} option to each question alongside mutually exclusive content options: selecting the reference answer is counted as correct; selecting \emph{Not Mentioned} indicates that the caption does not provide sufficient information to answer the question; and selecting an incorrect content option is counted as a hallucination. This constrained-choice protocol distinguishes correct recovery, missing evidence, and factual conflict at the question level, reducing subjective uncertainty in open-ended LLM scoring while improving stability and interpretability.
\setcounter{dbltopnumber}{2}
\section{Experiments}
\label{sec:experiments}

\subsection{Caption Evaluation}
\label{sec:existing-results}

To evaluate video selection, agentic data construction, and the resulting end-to-end Captioner, we use two complementary settings: (1) direct evaluation on established caption benchmarks, which measures detail coverage and factuality; and (2) cascade evaluation, in which a frozen caption supports downstream question answering, thereby measuring information completeness and task utility.

\begin{table}[H]
\centering
\small
\setlength{\tabcolsep}{2.5pt}
\captionof{table}{Detailed-captioning results on established benchmarks. Bold indicates the best open-source result per column. Published results are quoted; all others are reproduced using official protocols.}
\label{tab:direct-caption}
\begin{tabularx}{\textwidth}{l|*{2}{>{\centering\arraybackslash}X}|>{\centering\arraybackslash}X|*{5}{>{\centering\arraybackslash}X}}
\toprule
\multirow{2}{*}{\textbf{Model}} & \multicolumn{2}{c|}{\textbf{VDC Detailed}}
& \multicolumn{1}{c|}{\textbf{DREAM-1K}}
& \multicolumn{5}{c}{\textbf{Omni-Cloze}}\\
\cmidrule(lr){2-3}\cmidrule(lr){4-4}\cmidrule(lr){5-9}
& \multicolumn{1}{c}{\textbf{Acc. \%}} & \multicolumn{1}{c|}{\textbf{Score$\uparrow$}}
& \multicolumn{1}{c|}{\textbf{F1 score$\uparrow$}} & \multicolumn{1}{c}{\textbf{Visual \%}} & \multicolumn{1}{c}{\textbf{Audio \%}} & \multicolumn{1}{c}{\textbf{AV \%}} & \multicolumn{1}{c}{\textbf{Total \%}} & \multicolumn{1}{c}{\textbf{Score$\uparrow$}}\\
\midrule
\textbf{\textit{Proprietary Models}} & \multicolumn{2}{c|}{} & \multicolumn{1}{c|}{} & \multicolumn{5}{c}{}\\
GPT-4o \citep{openai2024gpt4o} & 46.3 & 2.5 & \multicolumn{1}{c|}{38.3} & 39.9 & 19.2 & 38.9 & 32.8 & 28.8\\
Gemini 3.5 Flash \citep{google2026gemini35flash} & 50.6 & 2.4 & \multicolumn{1}{c|}{42.0} & 49.7 & 25.1 & 51.2 & 41.6 & 36.1\\
Gemini 3.1 Pro \citep{google2026gemini31pro} & 51.0 & 2.5 & \multicolumn{1}{c|}{40.9} & 51.9 & 25.7 & 51.1 & 43.0 & 37.2\\
\midrule
\textbf{\textit{Open-Source Omni Models}} & \multicolumn{2}{c|}{} & \multicolumn{1}{c|}{} & \multicolumn{5}{c}{}\\
VideoLLaMA 2 \citep{cheng2024videollama2} & 38.1 & 1.9 & \multicolumn{1}{c|}{26.6} & 5.7 & 2.6 & 7.3 & 4.8 & 3.9\\
Qwen2.5-Omni \citep{xu2025qwen25omni} & 39.7 & 2.2 & \multicolumn{1}{c|}{31.6} & 10.4 & 12.9 & 18.9 & 12.9 & 11.0\\
Qwen3-Omni \citep{qwen2025omni} & 52.5 & 2.5 & \multicolumn{1}{c|}{35.9} & 47.4 & 40.4 & 49.7 & 45.3 & 33.4\\
MiniCPM-o-4.5 \citep{cui2026minicpmo45} & 50.8 & 2.5 & \multicolumn{1}{c|}{32.4} & 43.3 & 31.0 & 46.2 & 39.5 & 34.6\\
\midrule
\textbf{\textit{Open-Source Omni Caption Models}} & \multicolumn{2}{c|}{} & \multicolumn{1}{c|}{} & \multicolumn{5}{c}{}\\
OmniCaptioner-IF-3B \citep{ma2026omnicaptioner} & 46.9 & 2.3 & \multicolumn{1}{c|}{29.7} & 32.8 & 35.9 & 39.6 & 34.8 & 24.7\\
OmniCaptioner-IF-7B \citep{ma2026omnicaptioner} & 47.9 & 2.4 & \multicolumn{1}{c|}{31.7} & 31.2 & 29.6 & 40.8 & 32.0 & 28.3\\
AVoCaDO \citep{chen2025avocado} & 53.2 & 2.6 & \multicolumn{1}{c|}{35.9} & 42.1 & 46.4 & 46.6 & 44.2 & 40.9\\
video-SALMONN-2 \citep{tang2025videosalmonn2} & 52.0 & 2.6 & \multicolumn{1}{c|}{34.4} & 31.2 & 32.3 & 42.0 & 33.6 & 28.7\\
UGC-VideoCaptioner \citep{wu2025ugc} & 51.3 & 2.5 & \multicolumn{1}{c|}{31.2} & 36.0 & 26.2 & 40.6 & 33.3 & 29.0\\
\midrule
\captionermodel{} (Ours) & \textbf{57.9} & \textbf{2.9} & \multicolumn{1}{c|}{\textbf{36.2}} & \textbf{48.0} & \textbf{49.2} & \textbf{53.4} & \textbf{49.2} & \textbf{42.8}\\
\bottomrule
\end{tabularx}
\vspace{-6pt}
\end{table}

\noindent\textbf{Direct Evaluation.}\quad
We evaluate caption quality on VDC Detailed \citep{chai2025auroracap}, DREAM-1K \citep{wang2024tarsier}, and Omni-Cloze \citep{ma2026omnicaptioner}. VDC Detailed and DREAM-1K are visual-only captioning benchmarks: the former reports Accuracy and VDCscore through fine-grained visual question answering, while the latter reports the harmonic-mean F1 score to jointly measure descriptive accuracy and content completeness. Omni-Cloze evaluates the recoverability of visual, audio, and audiovisual information through constrained cloze questions, reporting Visual, Audio, AV, and Total Accuracy. Its Score rewards correct answers while penalizing incorrect non-\emph{Not Given} answers, thereby jointly reflecting correct recovery and hallucination errors.

As shown in Table~\ref{tab:direct-caption}, proprietary baselines include GPT-4o, Gemini 3.1 Pro, and Gemini 3.5 Flash. We further divide open-source models by intended use: open-source Omni models include VideoLLaMA 2, MiniCPM-o-4.5, Qwen2.5-Omni, and Qwen3-Omni; open-source Omni Caption models include video-SALMONN~2, OmniCaptioner, UGC-VideoCaptioner, AVoCaDO, and \captionermodel{}. \captionermodel{} obtains 57.9\% Accuracy and a 2.9 VDCscore on VDC Detailed, outperforming all proprietary, open-source Omni, and open-source Omni Caption baselines. It further achieves an F1 score of 36.2 on DREAM-1K and leads all open-source models on Omni-Cloze, demonstrating the strongest open-source performance across all three benchmarks.

\begin{table}[H]
\centering
\begin{minipage}[t]{0.49\textwidth}
\centering
\small
\setlength{\tabcolsep}{1.5pt}
\captionof{table}{Caption-to-QA cascade evaluation. The best result among open-source models in each column is shown in bold.}
\label{tab:cascade-caption}
\begin{tabularx}{\linewidth}{@{}l*{3}{>{\centering\arraybackslash}X}@{}}
\toprule
\textbf{Model} & \textbf{Daily-Omni} & \textbf{WorldSense} & \multicolumn{1}{c}{\textbf{Video-MME}}\\
\midrule
\multicolumn{4}{@{}l}{\textbf{\textit{Proprietary Models}}}\\
GPT-4o & 50.3 & 38.2 & 60.7\\
Gemini 3.5 Flash & 62.2 & 43.9 & 69.0\\
Gemini 3.1 Pro & 64.1 & 49.2 & 76.4\\
\midrule
\multicolumn{4}{@{}l}{\textbf{\textit{Open-Source Omni Models}}}\\
VideoLLaMA 2 & 57.6 & 41.5 & 62.8\\
Qwen2.5-Omni & 52.7 & 34.3 & 57.4\\
Qwen3-Omni & 58.5 & 38.8 & 63.6\\
MiniCPM-o-4.5 & 56.7 & 38.9 & 62.2\\
\midrule
\multicolumn{4}{@{}l}{\textbf{\textit{Open-Source Omni Caption Models}}}\\
OmniCaptioner-IF-3B & 54.7 & 37.0 & 58.9\\
OmniCaptioner-IF-7B & 53.5 & 39.1 & 60.7\\
AVoCaDO & 60.5 & 41.8 & 61.9\\
video-SALMONN-2 & 61.1 & 43.0 & 62.8\\
UGC-VideoCaptioner & 54.6 & 37.8 & 60.8\\
\midrule
\captionermodel{} & \textbf{64.6} & \textbf{45.7} & \textbf{64.8}\\
\bottomrule
\end{tabularx}
\end{minipage}
\hfill
\begin{minipage}[t]{0.49\textwidth}
\centering
\small
\setlength{\tabcolsep}{1.5pt}
\captionof{table}{Results on \bench{}. The best result among open-source models in each column is shown in bold.}
\label{tab:bench-results}
\begin{tabularx}{\linewidth}{@{}l*{3}{>{\centering\arraybackslash}X}@{}}
\toprule
\textbf{Model} & \textbf{Phys.} & \textbf{Non-Phys.} & \textbf{Total}\\
\midrule
\multicolumn{4}{@{}l}{\textbf{\textit{Proprietary Models}}}\\
GPT-4o & 36.3 & 20.4 & 26.1\\
Gemini 3.5 Flash & 37.3 & 21.3 & 27.0\\
Gemini 3.1 Pro & 47.8 & 39.1 & 42.3\\
\midrule
\multicolumn{4}{@{}l}{\textbf{\textit{Open-Source Omni Models}}}\\
VideoLLaMA 2 & 7.4 & 3.9 & 5.2\\
Qwen2.5-Omni & 17.3 & 10.9 & 13.2\\
Qwen3-Omni & 35.3 & 25.3 & 28.9\\
MiniCPM-o-4.5 & 34.7 & 26.7 & 29.5\\
\midrule
\multicolumn{4}{@{}l}{\textbf{\textit{Open-Source Omni Caption Models}}}\\
OmniCaptioner-IF-3B & 25.0 & 18.2 & 20.5\\
OmniCaptioner-IF-7B & 26.3 & 18.6 & 21.2\\
AVoCaDO & 38.1 & 40.9 & 39.4\\
video-SALMONN-2 & 31.6 & 25.4 & 27.3\\
UGC-VideoCaptioner & 24.8 & 17.0 & 19.9\\
\midrule
\captionermodel{} & \textbf{57.7} & \textbf{46.1} & \textbf{50.3}\\
\bottomrule
\end{tabularx}
\end{minipage}
\end{table}

\noindent\textbf{Cascade Evaluation.}\quad
We evaluate caption-to-QA performance on three omni-modal benchmarks: Daily-Omni \citep{zhou2025dailyomni}, WorldSense \citep{hong2026worldsense}, and Video-MME \citep{fu2025videomme}. This setting measures the fine-grained information retained by model- and system-generated captions. Using GPT-5.6 as a unified caption-only question-answering model, Table~\ref{tab:cascade-caption} reports Accuracy on the original questions. \captionermodel{} scores 64.6 on Daily-Omni, 0.5 percentage points above the strongest baseline, Gemini 3.1 Pro, and obtains 45.7 on WorldSense and 64.8 on Video-MME. Although proprietary models such as Gemini 3.1 Pro remain ahead on the latter two benchmarks, \captionermodel{} ranks first among all compared open-source methods.

\subsection{Results and Analysis of \bench{}}
\label{sec:bench-results}

\noindent\textbf{Model Performance.}\quad
Table~\ref{tab:bench-results} reports Caption-QA Accuracy on \bench{} for all questions (Total), the physics-question subset (Phys.), and the non-physics subset (Non-Phys.). \captionermodel{} achieves 50.3\%, 57.7\%, and 46.1\% on the three metrics, respectively, outperforming all proprietary and open-source baselines. On Total and Phys., \captionermodel{} improves over the strongest baseline, Gemini 3.1 Pro, by 8.0 and 9.9 percentage points, respectively; on Non-Phys., it surpasses the strongest baseline, AVoCaDO, by 5.3 percentage points. The more pronounced gain on the physics subset indicates that \ppm{}-enriched Agent supervision enables the Captioner to retain more physical evidence about object properties, interaction relations, and state changes. Meanwhile, the concurrent improvement on Non-Phys. shows that this enhancement in physical perception does not come at the expense of general audiovisual information; instead, cross-tool evidence aggregation improves the overall information completeness and cross-modal recoverability of the caption.

\begin{table}[t]
\centering
\begin{minipage}{\textwidth}
\centering
\includegraphics[width=\textwidth]{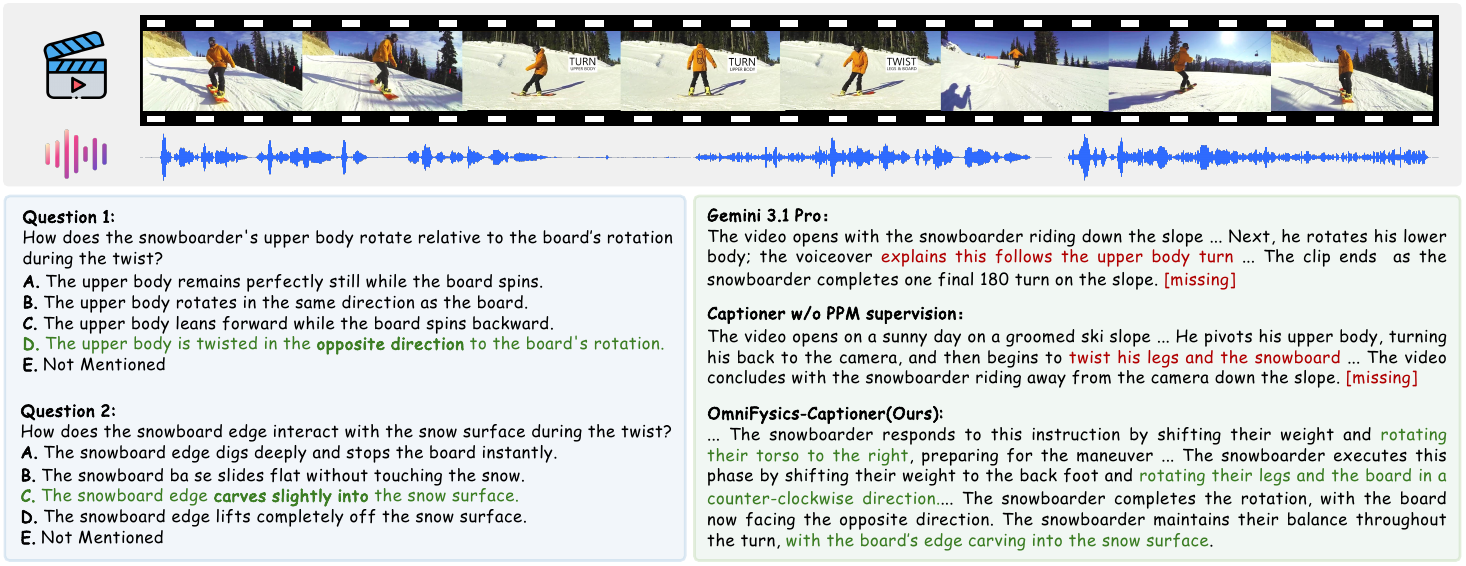}
\captionof{figure}{Case study. Gemini 3.1 Pro and the Captioner without \ppm{} supervision omit key physical details, while physical perception enables \captionermodel{} to achieve stronger omni-modal understanding.}
\label{fig:case-study}
\end{minipage}
\vspace{8pt}

\captionof{table}{Three-level analysis of \agent{} through one-pass baselines, an online \ppm{} ablation, and transfer of \ppm{}-enriched supervision to a tool-free Captioner.}
\label{tab:agent-phys-ablation}
\begin{minipage}{0.70\columnwidth}
\centering
\small
\setlength{\tabcolsep}{1.5pt}
\begin{tabularx}{\linewidth}{@{}>{\raggedright\arraybackslash}X*{3}{>{\centering\arraybackslash}p{0.19\linewidth}}@{}}
\toprule
\textbf{Method} & \textbf{Daily-Omni} & \textbf{OPC Total} & \textbf{OPC Phys.}\\
\midrule
\multicolumn{4}{@{}l}{\textbf{\textit{One-Pass Captioning}}}\\
\quad Qwen3-Omni & 58.5 & 28.9 & 35.3\\
\quad Gemini 3.1 Flash-Lite & 60.6 & 29.9 & 41.5\\
\midrule
\multicolumn{4}{@{}l}{\textbf{\textit{Online Agent Inference}}}\\
\quad Full toolbox & 72.6 & 53.0 & 63.6\\
\quad w/o \ppm{} & 70.2 & 51.0 & 55.8\\
\midrule
\multicolumn{4}{@{}l}{\textbf{\textit{Tool-Free Captioner}}}\\
\quad \ppm{}-enriched sup. & 64.6 & 50.3 & 57.7\\
\quad no-\ppm{} sup. & 62.7 & 45.7 & 53.1\\
\bottomrule
\end{tabularx}
\end{minipage}
\end{table}

\subsection{Multi-Level Analysis of \agent{}}
\label{sec:agent-analysis}

This section examines \agent{} at three complementary levels: the system-level effectiveness of active evidence acquisition, the component-level contribution of \ppm{} during caption construction, and the deployment-level transfer of \ppm{}-enriched supervision to a Captioner. We evaluate on two representative benchmarks using the setup described under Caption Evaluation throughout. Figure~\ref{fig:case-study} complements the quantitative analysis with a representative comparison of physical details retained by different captions.

\noindent\textbf{Effectiveness of Active Evidence Acquisition.}\quad
Physical interactions and state changes are often localized in short temporal intervals and may be missed by one-pass observation. We first examine whether the active acquisition process described under \agent{} improves the amount of information retained in the final caption. Qwen3-Omni and Gemini 3.1 Flash-Lite serve as the system's primary tool and Planner, respectively. We compare their standalone captioning performance with that of the complete \agent{} system built upon these components. As shown in Table~\ref{tab:agent-phys-ablation}, \agent{} obtains 72.6 on Daily-Omni, 53.0 on \bench{} Total, and 63.6 on \bench{} Phys. It exceeds Qwen3-Omni by 14.1, 24.1, and 28.3 points, respectively, and Gemini 3.1 Flash-Lite by 12.0, 23.1, and 22.1 points. These gains show that the complete active evidence-acquisition workflow produces captions with more recoverable information than either component used alone for one-pass captioning. Across fine-grained captioning and cascade evaluations, the workflow integrates temporally distributed and modality-specific evidence through iterative localized observation and cross-tool aggregation, thereby yielding captions with substantially broader recoverable coverage of audiovisual and physical details.

\noindent\textbf{Contribution of \ppm{}.}\quad
The preceding comparison validates the complete Agent but does not determine whether the dedicated physical Observer contributes beyond the general audio and visual tools. We therefore construct a matched ablation in which only \ppm{} is removed, while the Planner, Finalizer, ALM/VLM access, call budget, caption instruction, and decoding settings remain unchanged. Removing \ppm{} decreases Daily-Omni from 72.6 to 70.2 and \bench{} Total from 53.0 to 51.0. More notably, \bench{} Phys. decreases from 63.6 to 55.8, a 7.8-point drop. The substantially larger degradation on the physics-specific subset indicates that \ppm{} contributes targeted evidence about object properties, contact relations, material responses, and state transitions, rather than producing an undifferentiated increase in caption detail.

\noindent\textbf{Transfer of \ppm{}-Enriched Supervision.}\quad
The practical purpose of Agent-generated supervision is to transfer evidence-rich captioning behavior to an end-to-end model that does not require online tools. To test this transfer, we train two Captioners on paired captions generated from the same video manifest. The models use the same initialization and training configuration; the only difference in constructing the two training datasets is whether \ppm{} is used as an Observer when generating the supervision captions. As reported in Table~\ref{tab:agent-phys-ablation}, \ppm{}-enriched supervision improves Daily-Omni from 62.7 to 64.6, \bench{} Total from 45.7 to 50.3, and \bench{} Phys. from 53.1 to 57.7. These results further demonstrate that physical evidence acquired by \agent{} can be partially transferred to single-pass, tool-free caption generation. They also highlight the value of physical-perception tools in omni-modal captioning, particularly for physical perception, and further show that physical perception can strengthen omni-modal understanding.

\section{Conclusion}
\label{sec:conclusion}

This work presents a unified framework for physics-aware omni-modal captioning. We construct Daily-Physics 50K from physics-rich videos, use \agent{} to acquire temporally localized audiovisual and physical evidence, and train \captionermodel{} for tool-free inference. We also introduce \bench{} to evaluate the recovery of physical and cross-modal information from generated captions. Experiments show state-of-the-art open-source performance on detailed audiovisual captioning benchmarks. Multi-level analyses demonstrate that active evidence acquisition improves information coverage, that the \ppm{} Observer provides targeted cues about object properties, interactions, material responses, and state transitions, and that \ppm{}-enriched supervision transfers these gains to the end-to-end Captioner.

For embodied intelligence, robotic manipulation, and world models, physical perception bridges multimodal observation and reliable understanding of dynamic environments. By providing scalable physics-rich supervision and evaluation, our framework offers a foundation for building such systems. We will release Daily-Physics 50K, \captionermodel{}, and \bench{} to support research on physical perception and omni-modal captioning.

%% file: appendix_content.tex
\section{Video Dataset Details}
\label{app:corpus}
\label{app:taxonomy}

\subsection{Source Dataset Descriptions}
\label{app:source-datasets}

The videos used to construct our training dataset come from multiple sources to ensure diverse audiovisual content. Below, we summarize the scale, content, and annotation tasks of each source dataset according to its original paper or official dataset page.

\noindent\textbf{AVQA.}\quad
MUSIC-AVQA~\citep{li2022musicavqa} studies questions about visual objects, sounds, and their associations in dynamic audio-visual scenes. It contains 9,288 musical-performance videos totaling more than 150 hours and 45,867 question--answer pairs generated from 33 templates. The questions span three modal scenarios and nine question types, requiring multisensory perception and spatiotemporal reasoning.

\noindent\textbf{Bilibili-Videos.}\quad
Bilibili-Videos consists of user-generated videos from the Bilibili platform rather than an academic benchmark with a fixed task and official split. Its content includes everyday demonstrations, sports, entertainment, and uploads containing multiple consecutive events, with diverse scenes, editing styles, and activity types. Use and redistribution remain subject to the original licenses and platform terms.

\noindent\textbf{EPIC-Kitchens.}\quad
EPIC-KITCHENS-100~\citep{damen2022epickitchens100} provides 100 hours of unscripted egocentric audio-visual recordings captured with head-mounted cameras in 45 kitchens across four cities, with approximately 90K action segments. Its close-range manipulation sequences include contact, tool use, pouring, cutting, heating, and before--after state changes, together with verb and noun annotations for action recognition, action detection, and action anticipation.

\noindent\textbf{LLaVA-Video.}\quad
LLaVA-Video-178K~\citep{zhang2025llavavideo} was constructed for video instruction tuning and contains 178,510 caption entries, 960,792 open-ended question--answer items, and 196,198 multiple-choice question--answer items. It covers academic and YouTube videos across multiple duration ranges and organizes synthetic annotations into detailed captioning, open-ended question answering, and multiple-choice question answering tasks.

\noindent\textbf{LU-AVS.}\quad
LU-AVS~\citep{liu2024luavs} benchmarks audio-visual segmentation in long untrimmed videos, with precise sounding intervals and dense spatial annotations. The original release reports 10M masks across 6.6K videos and 11M bounding boxes across 7K videos, with substantially longer videos and more silence than trimmed audio-visual segmentation datasets. It supports evaluation of intermittent sound, sounding-object localization, and long-range audio-visual changes.

\noindent\textbf{VideoMind.}\quad
VideoMind~\citep{yang2025videomind} contains 103K audio-equipped videos with hierarchical factual, abstract, and intent descriptions; 3K manually validated samples are reserved by its authors for evaluation. The factual layer describes subjects, places, times, events, and actions, the abstract layer summarizes semantics across segments, and the intent layer captures deeper purposes in the context of the complete video.

\noindent\textbf{WISA.}\quad
WISA-32K~\citep{wang2025wisa} was collected for physics-aware text-to-video generation and organizes 32K videos around 17 physical laws in dynamics, thermodynamics, and optics. Its observable processes include motion and collision, fluid behavior, heat transfer and phase change, and optical phenomena such as reflection and refraction, accompanied by text descriptions grounded in physical laws.

\noindent\textbf{FineVideo.}\quad
FineVideo~\citep{huggingface2024finevideo} is a Hugging Face collection of 43,751 Creative Commons YouTube videos with provenance metadata and rich time-aware annotations, including scenes, activities, audio-visual correlation, narrative progression, and editing details. It primarily contains long-form, open-domain videos in which an individual video may include multiple scenes and consecutive events. Licensing, attribution, and removal requirements are specified on the dataset page.

\noindent\textbf{unAV-100.}\quad
unAV-100~\citep{geng2023unav100} targets dense localization of audio-visual events in untrimmed video. It contains 10K videos and more than 30K events across 100 categories, with 2.8 audio-visual events per video on average and possible temporal overlap. The benchmark evaluates event classification and temporal-boundary localization in complex scenes.

\subsection{Event Queries and Clip Construction}
\label{app:event-queries}

Using the nine sources above, we define 23 observable physical-dynamic event subcategories, organized into the following six top-level physical-event categories:

\begingroup
\setlength{\parindent}{1.5em}
\setlength{\parskip}{2pt}

\textbf{Sports biomechanics:} \emph{1. Contact and collision in sports; 2. Fluid dynamics in water sports; and 3. Running, jumping, and aerial motion.}

\textbf{Motion and trajectories:} \emph{4. Projectile--air interaction; 5. Chain reactions and transmission; and 6. Rotation and balance.}

\textbf{Impact and material response:} \emph{7. Collision and impact; 8. Momentum transfer; 9. Swinging and oscillation; and 10. Fracture and deformation.}

\textbf{Everyday fluid dynamics:} \emph{11. Splashes and droplets; 12. Flow, mixing, and diffusion; and 13. Surface tension, cleaning, and special fluids.}

\textbf{Household thermodynamics and optics:} \emph{14. Flames and sparks; 15. Melting, freezing, evaporation, and condensation; 16. Refraction and reflection; 17. Everyday mechanical actions; 18. Fluid mixing and dissolution; 19. Household airflow, suction, and pressure; and 20. Kitchen heat and phase changes.}

\textbf{Surface friction and contact:} \emph{21. Friction and sliding; 22. Adhesion, peeling, and contact separation; and 23. Braking, frictional stopping, and grinding.}
\endgroup

Each event-subcategory query contains an event name, a phenomenon definition, representative scenes, and Chinese and English keywords. The retrieval and screening procedure is implemented as follows.

\noindent\textbf{Frame retrieval and CATA segmentation.}\quad
We first sample every video at 1 fps and use Qwen3-VL-Embedding-8B to encode the 23 event queries and sampled frames. Query vectors are computed once per retrieval run, while each frame vector is reused for all queries. Cosine similarity between normalized frame and query vectors assigns a candidate event category to each retrieved frame. CATA averages the frame-level similarities in a 1-second sliding window and then merges candidate frames in temporal order. A segment continues only when the event category remains unchanged and the next candidate is no more than 1 second away; a category change or a gap longer than 1 second starts a new segment. Segments shorter than 3 seconds are removed. Short-process clips are capped at 60 seconds and longer continuous responses are split over time. A separate long-process branch keeps 60--600-second videos for multi-step operations, long causal chains, and slowly changing states.

\noindent\textbf{Human quality audit.}\quad
We draw a category-stratified random sample of 500 clips from the final training manifest, covering all six top-level physical-event categories, and ask human annotators to watch the complete audiovisual input and answer two binary questions for the same set of clips. The \emph{temporal-completeness} question asks whether the beginning, interaction, and visible outcome of the main action all fall inside the clip, without truncation by the clip boundary or disruptive editing. The \emph{category-consistency} question asks whether the assigned physical event is directly observable and agrees with the assigned category, without relying on titles, narration, or an unobserved cause. The two pass rates are aggregated independently from the collected binary judgments. The audit finds that 98.2\% of clips are temporally complete and 96.9\% contain physical events consistent with their assigned categories. These results show that the retrieval, segmentation, and screening pipeline preserves a complete process and the intended category for the large majority of clips, while leaving a small residual set with incomplete boundaries or ambiguous multi-event semantics.

The high pass rates compare favorably with the quality and computational profiles of existing video-data construction pipelines. WISA-32K manually collects physics videos, applies shot detection and aesthetic filtering, generates captions with Qwen2-VL, and performs five rounds of qualitative and three rounds of quantitative physical annotation with GPT-4o mini \citep{wang2025wisa}. At a larger scale, Panda-70M generates eight candidate captions with cross-modal teacher models and applies a ninth learned retrieval model for annotation selection \citep{chen2024panda70m}. ShareGPT4Video first filters videos through generated semantic descriptions, extracts nonredundant keyframes, invokes GPT-4V over successive keyframe pairs, aggregates the differential descriptions with GPT-4, and finally conducts manual quality inspection \citep{chen2024sharegpt4video}.

CATA removes these generative and manual operations from the candidate-discovery stage. The 23 event-query embeddings are cached, every sampled-frame embedding is reused across all queries, and clip boundaries are formed by a single linear scan over frame responses. Generative annotation and human inspection can therefore focus on the retrieved physical-event clips instead of the complete heterogeneous video pool. The resulting 98.2\% temporal completeness and 96.9\% category consistency demonstrate that this retrieval-first design substantially reduces the number of expensive model calls and manual screening decisions required before caption generation, while preserving highly reliable physical-event content and temporal boundaries.

\FloatBarrier
\section{Agent-Based Caption Generation}
\label{app:agent-details}

\subsection{Multimodal Evidence Acquisition}
\label{app:evidence-acquisition}

Our \agent{} pipeline coordinates multiple multimodal models to collect complementary evidence for detailed caption generation. Gemini 3.1 Flash-Lite serves as the Planner: it uses a lightweight audiovisual proxy to establish a global event timeline, locates intervals that require closer inspection, and selects the appropriate Observer. The Observer toolbox follows the modality-specific roles defined in the main paper. Qwen3-Omni acts as the ALM for speech, music, and environmental sounds; Qwen3.6-35B-A3B acts as the VLM for local visual events, text, and visible state transitions; and the Physical Perception Model (\ppm{}) examines representative frames for material, contact, support, deformation, motion, and likely physical outcomes.

Evidence acquisition proceeds from coarse temporal navigation to focused verification. In the first round, the Planner reads the audiovisual proxy to build a cross-modal event skeleton that summarizes the initial state, key events and transitions, and visible outcomes. It then identifies information-dense intervals and issues a batch of modality-directed questions. For each admitted request, the Executor crops the corresponding interval from the original audiovisual stream and routes the local media to the selected Observer. Requests in the same round are executed in parallel, and their source-attributed outputs are written to Evidence Memory. In later rounds, the Planner reviews this accumulated evidence, identifies missing or uncertain details, and narrows the modality, temporal window, and question focus for follow-up observations. The active loop ends when the evidence is sufficient or no useful follow-up remains.

\subsection{Coverage Completion and Caption Synthesis}
\label{app:coverage-synthesis}

After the active loop exits, Coverage performs a final deterministic pass without invoking the Planner. Based on the video duration, available modalities, and observations already stored in Evidence Memory, it adds only the baseline evidence that remains under-covered: global audio context when needed, visual checks distributed across the timeline, and representative-frame analysis for salient physical interactions that have not been adequately verified. These requests follow the same Execute--Observe--Update path and are appended to the shared memory. Coverage therefore closes residual modality and temporal gaps while preserving the active loop as the primary mechanism for targeted investigation.

Gemini 3.1 Pro serves as the Finalizer. It makes no further media or tool calls, but organizes the final Evidence Memory and the retained non-media planning context into a temporally coherent, nonredundant caption, preserving uncertainty when observations cannot be reconciled. These captions are frozen as supervision for the end-to-end Captioner and are evaluated using the caption and caption-to-QA protocols described in the main paper.
\FloatBarrier

\section{Implementation Details}
\label{app:model-details}
\label{app:phys-details}

\subsection{Physical Perception Model (PPM)}
\label{app:ppm-details}
The Physical Perception Model (\ppm{}) serves as a dedicated Observer in \agent{}, extracting object-level physical evidence from representative video frames, including material properties, contact and support relations, deformation, motion, and likely state changes. To train this capability, we use GPT-5.4 to construct approximately 2M Chinese and English question--answer examples based on Open Images V7. We obtain \ppm{} by full-parameter supervised fine-tuning of Qwen3.6-35B-A3B on these examples.

\ppm{} identifies salient objects and uses approximate normalized bounding boxes as spatial anchors for associating physical properties with specific objects. It maintains object references across single- and multi-turn queries, compares physical properties between objects, and performs qualitative reasoning about sliding, stability, deformation, interaction outcomes, and changes caused by different materials or external conditions. The bounding boxes establish object--evidence correspondence rather than serving as an independent localization objective. Within the Agent pipeline, \ppm{} converts representative frames into concise object-centric physical evidence that complements audio and visual observations with properties, interaction relations, and state-change cues.

\subsection{Training Configuration}
\label{app:training-configuration}
Table~\ref{tab:model-training-details} lists the key training hyperparameters for the \ppm{} Observer and \captionermodel{}.

\begin{table}[H]
\centering
\caption{Training hyperparameters for the \ppm{} Observer and \captionermodel{}.}
\label{tab:model-training-details}
\begin{minipage}{0.60\columnwidth}
\centering
\footnotesize
\setlength{\tabcolsep}{3pt}
\begin{tabularx}{\linewidth}{@{}p{0.23\linewidth}>{\centering\arraybackslash}p{0.33\linewidth}>{\centering\arraybackslash}X@{}}
\toprule
\textbf{Configuration} & \textbf{\ppm{} Observer} & \textbf{\captionermodel{}}\\
\midrule
Training data & 2M & Daily-Physics 50K\\
GPU usage & 32 $\times$ NVIDIA H100 & 16 $\times$ NVIDIA H100\\
GBS & 64 & 16\\
Epochs & 1 & 2\\
Peak LR & $5\times10^{-6}$ & $5\times10^{-6}$\\
LR schedule & Cosine & Cosine\\
\bottomrule
\end{tabularx}
\end{minipage}
\end{table}

The \ppm{} Observer takes a single image as input and is trained for one epoch with full-parameter supervised fine-tuning (SFT). \captionermodel{} is trained for two epochs on the frozen Daily-Physics 50K video--caption pairs generated by the Agent pipeline. During training and evaluation, videos are sampled at 2 fps with audio enabled, and up to 256 frames are retained. We use the checkpoint from the end of the second epoch for all reported evaluations.

\section{OmniPhysCap (OPC) Benchmark Details}
\label{app:bench}

\subsection{Independent Evaluation-Video Curation}
\label{app:opc-curation}
\bench{} is constructed from an independently collected evaluation pool that is completely separate from Daily-Physics 50K. No benchmark clip, source video, or temporally overlapping segment appears in the Captioner training corpus. We further remove exact and perceptual near-duplicates across the two sets using source identity, file hashes, and multi-frame visual similarity. This strict separation provides a consistent held-out evaluation setting for \captionermodel{} and all other evaluated models, supporting a fair comparison across systems.

Candidate clips first undergo media-level validation to remove unreadable, incomplete, or otherwise unusable audiovisual files. Gemini 3.1 Pro then reviews each complete clip along three dimensions: (i)~\emph{media quality}, covering visual clarity and usable audio; (ii)~\emph{event quality}, covering temporal coherence, physical-process observability, and event completeness; and (iii)~\emph{annotation quality}, covering ambiguity and whether the available evidence supports reliable annotation. This screening removes nearly static footage, repetitive actions without observable outcomes, routine interface or gameplay content, decorative animation, text- or narration-dominant videos, and clips whose key action or result falls outside the temporal boundary. Accordingly, aesthetic quality is operationalized through visibility, temporal coherence, and annotatability rather than production polish or novelty.

The remaining videos are deduplicated and grouped by visual scene and event semantics to reduce repeated templates and highly similar recordings. We then select 1,000 clips with balanced coverage of physical events, durations, source domains, and audiovisual content. The selected videos remain isolated from Daily-Physics 50K throughout benchmark construction.

\subsection{Evidence-Grounded Construction}
\label{app:opc-construction}
For each candidate video, the final \agent{} caption and successful audio, visual, and physical observations are organized into an evidence dossier. A reasoning model uses this dossier to propose typed questions, reference answers, and mutually exclusive distractors, but the Agent evidence does not determine the final gold answer. Gemini 3.1 Pro independently reads the complete audiovisual video and performs consistency checks on answer support, distractor exclusivity, temporal scope, and wording. Candidates that are ambiguous, unsupported, or internally inconsistent are discarded or revised.

Verified candidates are semantically deduplicated and selected to increase coverage of question types, event categories, and modality-specific information. Each video retains 5--10 questions, and human annotators cross-validate every retained item against the complete audiovisual input, removing questions with uncertain evidence, overlapping options, or inappropriate temporal scope. The final benchmark contains 8,000 questions over 1,000 videos. General questions assess global scene, object, and action understanding; Physics Interaction questions examine contact, support, collision, deformation, and motion relations; and Physics Outcome questions target observable state changes and consequences. Audio, AV Alignment, and Speech/OCR questions respectively evaluate acoustic evidence, temporal correspondence between sound and vision, and spoken or written information. Temporal questions test event ordering and duration, while Ending/Secondary questions probe final states and less salient events outside the dominant action. This decomposition enables \bench{} to diagnose physical-process coverage, cross-modal grounding, temporal understanding, and information omissions that can be obscured by a single aggregate caption score.

\begin{table}[!t]
\centering
\caption{Question-type distribution in \bench{}. The two physics types contain 2,938 questions in total.}
\label{tab:opc-question-types}
\begin{minipage}{0.65\columnwidth}
\centering
\small
\setlength{\tabcolsep}{4pt}
\begin{tabularx}{\linewidth}{@{}>{\raggedright\arraybackslash}X*{2}{>{\centering\arraybackslash}p{0.22\linewidth}}@{}}
\toprule
\textbf{Question type} & \textbf{Questions} & \textbf{Percentage (\%)} \\
\midrule
General & 1,704 & 21.3 \\
Physics Interaction & 1,573 & 19.7 \\
Physics Outcome & 1,365 & 17.1 \\
Audio & 835 & 10.4 \\
Ending/Secondary & 686 & 8.6 \\
Speech/OCR & 660 & 8.3 \\
AV Alignment & 625 & 7.8 \\
Temporal & 552 & 6.9 \\
\bottomrule
\end{tabularx}
\end{minipage}
\end{table}

\subsection{Caption-to-QA Evaluation Protocol}
\label{app:opc-evaluation}
Each evaluated system generates one caption for every benchmark video. GPT-5.6 is used as a fixed caption-only Judge and receives only the generated caption, a question, and five answer options: four mutually exclusive content options and \emph{Not Mentioned}. It does not access the video, construction evidence, question category, or gold answer. Selecting the reference option is counted as Correct, while selecting \emph{Not Mentioned} indicates that the caption omits the required evidence. The primary Coverage metric is computed per video and macro-averaged; Correct and Not Mentioned are additionally aggregated over all questions. Category Coverage is computed over videos containing the corresponding question family.

Table~\ref{tab:opc-detailed-results} reports question-level results computed from the same frozen captions and GPT-5.6 Judge outputs used for the OPC results in the main paper. Correct measures whether the caption supports the reference answer, while \emph{Not Mentioned} captures missing evidence. Both columns are question-level micro averages over all 8K questions, whereas the Total score in the main paper is video-level macro-averaged Coverage.

\begin{table}[H]
\centering
\caption{Question-level Correct and Not Mentioned rates on \bench{} (\%) over all 8K questions, computed from the same evaluation outputs as the main-paper OPC results. Correct is question-level micro accuracy; the main-paper Total is video-level macro-averaged Coverage.}
\label{tab:opc-detailed-results}
\begin{minipage}{0.65\columnwidth}
\centering
\footnotesize
\setlength{\tabcolsep}{2pt}
\begin{tabularx}{\linewidth}{@{}>{\raggedright\arraybackslash}X*{2}{>{\centering\arraybackslash}p{0.22\linewidth}}@{}}
\toprule
\textbf{Model} & \textbf{Correct} & \textbf{Not Ment.} \\
\midrule
\multicolumn{3}{@{}l}{\textbf{\textit{Proprietary Models}}}\\
GPT-4o & 26.2 & 71.3 \\
Gemini 3.5 Flash & 27.1 & 70.9 \\
Gemini 3.1 Pro & 42.6 & 53.9 \\
\midrule
\multicolumn{3}{@{}l}{\textbf{\textit{Open-Source Omni Models}}}\\
VideoLLaMA 2 & 5.2 & 93.6 \\
Qwen2.5-Omni & 13.4 & 85.2 \\
Qwen3-Omni & 29.0 & 68.6 \\
MiniCPM-o-4.5 & 29.6 & 67.7 \\
\midrule
\multicolumn{3}{@{}l}{\textbf{\textit{Open-Source Omni Caption Models}}}\\
OmniCaptioner-IF-3B & 20.5 & 76.2 \\
OmniCaptioner-IF-7B & 21.4 & 75.2 \\
AVoCaDO & 39.2 & 56.8 \\
video-SALMONN-2 & 27.5 & 69.4 \\
UGC-VideoCaptioner & 19.9 & 77.9 \\
\midrule
\captionermodel{} & \textbf{50.9} & \textbf{44.4} \\
\bottomrule
\end{tabularx}
\end{minipage}
\end{table}

\subsection{Human Validation of OPC}
\label{app:opc-human-validation}
To quantify question quality beyond model-based verification, we audit questions drawn from 200 benchmark videos. Human reviewers inspect the complete audiovisual input and check whether each question is answerable from observable evidence, whether the reference answer is supported, whether the content options are mutually exclusive, and whether the wording and temporal scope are unambiguous. An audited question is counted as reliable only when these requirements are jointly satisfied. The resulting question-reliability rate is 95.1\%, indicating that the large majority of audited questions provide a clear and uniquely supported evaluation target.

We separately assess the reliability of the fixed GPT-5.6 Judge under the same caption-only decision setting used for benchmark scoring. Human reviewers receive the frozen caption, question, and five answer options, without access to the original audiovisual input, construction evidence, or protected gold answer. Agreement is measured by whether the GPT Judge and human reviewer assign the same scoring decision to an audited case. Across cases drawn from the same 200-video subset, the raw GPT--human agreement reaches 97.4\%. Together with the 95.1\% question-reliability rate, this result supports the use of the fixed Judge for scalable OPC evaluation while retaining human review as an independent check on benchmark-item quality and automated scoring.

\subsection{Qualitative Case Studies}
\label{app:opc-case-studies}
\begin{figure*}[p]
\centering
\includegraphics[width=\textwidth]{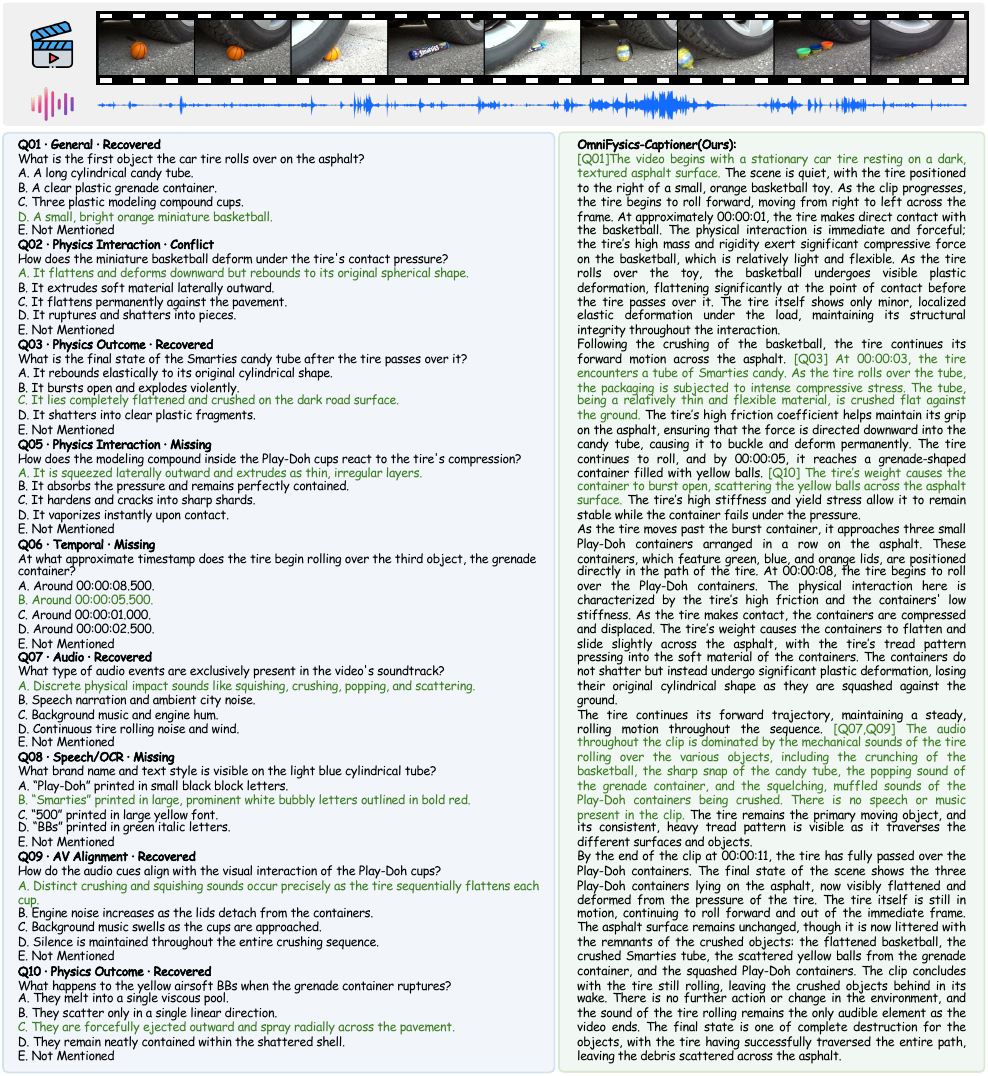}
\caption{Full \bench{} case study for tire crushing. The left panel contains all nine questions and Gold answers; the right panel shows the complete frozen caption with the original evidence excerpts supporting the five recovered answers. The case jointly probes object order, deformation, rupture, temporal localization, audio, OCR, and audio--visual alignment.}
\label{fig:opc-case-tire}
\end{figure*}

\begin{figure*}[p]
\centering
\includegraphics[width=\textwidth]{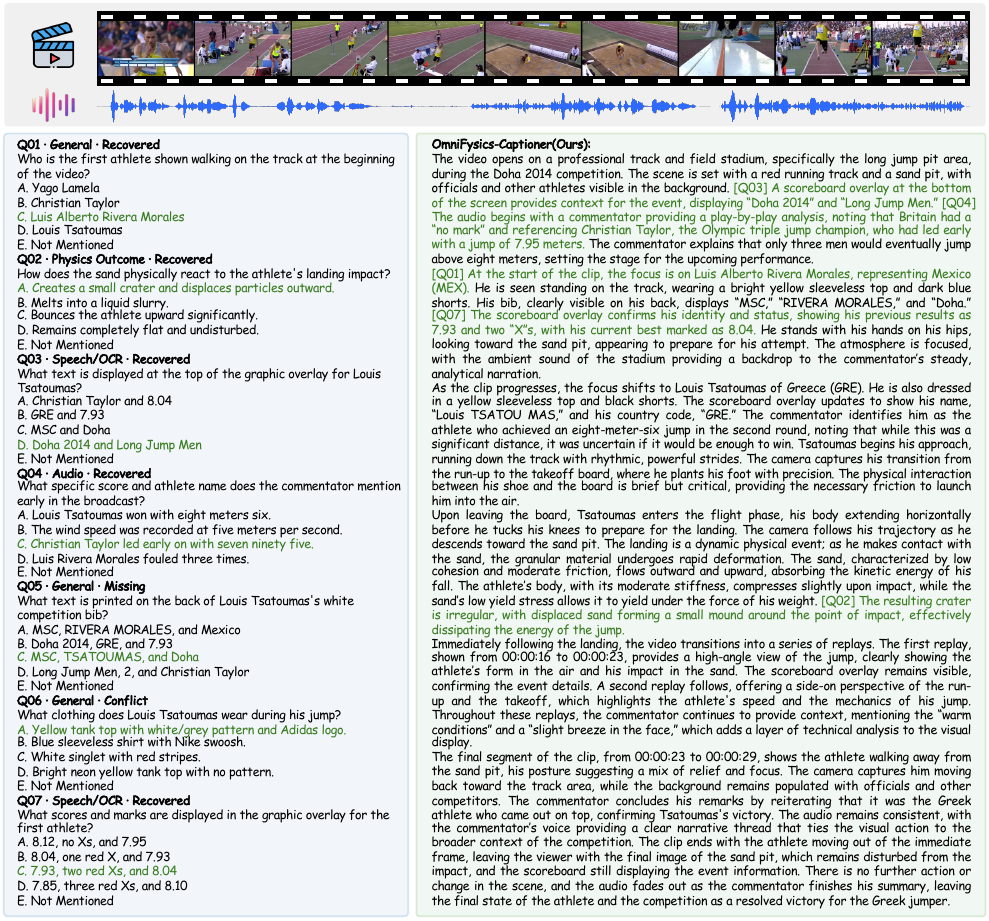}
\caption{Full \bench{} case study for long jump. The seven questions test athlete identity, granular landing impact, commentary, clothing, and on-screen competition information. Green caption excerpts provide the exact textual evidence used to recover five reference answers.}
\label{fig:opc-case-long-jump}
\end{figure*}

Figures~\ref{fig:opc-case-tire} and~\ref{fig:opc-case-long-jump} illustrate how \bench{} connects temporally localized audiovisual evidence to its complete set of diagnostic questions. Each case lists every question, all answer options, and the Gold answer on the left, while the right panel presents the frozen \captionermodel{} caption in full. Green excerpts mark the original caption spans from which the fixed caption-only Reader recovers the reference answer, while \emph{Missing} badges identify evidence omissions. The two examples cover sequential deformation and rupture, audio--visual event alignment, granular impact, broadcast speech, and on-screen results.

\section{Evaluation Details}
\label{app:evaluation}

\subsection{Benchmark Overview}
\label{app:benchmark-overview}
\noindent\textbf{VDC Detailed.}\quad
VDC Detailed~\citep{chai2025auroracap} contains 1,027 open-domain videos paired with manually inspected, structured reference descriptions averaging approximately 501 words. The references cover objective facts, background context, camera behavior, and detailed event timelines. Its VDCscore decomposes each reference into standardized short question--answer probes and tests whether the predicted caption preserves the corresponding information, making it suitable for evaluating long, information-dense descriptions. As reported in the main paper, \captionermodel{} achieves 57.9\% accuracy and a VDCscore of 2.9, the strongest result among the evaluated models.

\noindent\textbf{DREAM-1K.}\quad
DREAM-1K~\citep{wang2024tarsier} comprises 1,000 clips balanced across live-action films, animated films, stock footage, YouTube videos, and short-form videos. Each clip contains dynamic events that cannot be reliably inferred from a single frame. Its AutoDQ protocol extracts atomic events from reference and generated descriptions, then applies bidirectional entailment to compute precision, recall, and their harmonic-mean F1 score. This formulation jointly measures event coverage and unsupported content; \captionermodel{} obtains an F1 score of 36.2, ranking first among the compared open-source models.

\noindent\textbf{Omni-Cloze.}\quad
Omni-Cloze~\citep{ma2026omnicaptioner} evaluates detailed perception in visual-only, audio-only, and audio--visual settings. It contains 2,320 clips and 69,600 human-validated cloze blanks across 9 domains and 47 subcategories, with a fixed set of 30 blanks per clip. Each blank includes a \emph{Not Given} option, allowing the protocol to separate omitted information from an incorrect asserted detail. \captionermodel{} achieves 49.2\% overall accuracy and a score of 42.8, with its strongest modality split on audio--visual questions at 53.4\%, and leads all compared open-source models across the reported metrics.

\noindent\textbf{Daily-Omni.}\quad
Daily-Omni~\citep{zhou2025dailyomni} targets temporally aligned audio--visual reasoning in real-life scenarios. Its 684 videos and 1,197 multiple-choice questions cover six task families: AV event alignment, event sequence, reasoning, inference, comparison, and context understanding. The benchmark therefore tests whether a caption preserves not only modality-specific events but also their temporal correspondence and cross-modal dependencies. \captionermodel{} reaches 64.6\% caption-to-QA accuracy, the highest result in the comparison and 0.5 percentage points above Gemini 3.1 Pro.

\noindent\textbf{WorldSense.}\quad
WorldSense~\citep{hong2026worldsense} assesses real-world omni-modal recognition, understanding, and reasoning. It contains 1,662 synchronized audio--visual clips spanning 8 domains and 67 subcategories, together with 3,173 multiple-choice questions organized into 26 tasks. Its questions emphasize coupled audio--visual evidence across speech, environmental sounds, and music, rather than independent single-modality recognition. \captionermodel{} obtains 45.7\% accuracy, leading all compared open-source models and approaching Gemini 3.1 Pro at 49.2\%.

\noindent\textbf{Video-MME.}\quad
Video-MME~\citep{fu2025videomme} provides broad video-understanding coverage over 900 videos and 2,700 multiple-choice questions from 6 domains and 30 fine-grained categories. Videos are divided into short ($<2$ minutes), medium (4--15 minutes), and long (30--60 minutes) groups, and the benchmark includes audio and available subtitles in addition to visual content. This design evaluates perception, reasoning, information synthesis, and robustness across substantially different temporal contexts. Under the unified caption-to-QA protocol, \captionermodel{} achieves 64.8\% accuracy and ranks first among the evaluated open-source models.

\subsection{Standard Benchmark Evaluation}
\label{app:standard-evaluation}
VDC Detailed is evaluated with its public caption-to-short-answer pipeline and reports both answer accuracy and VDCscore. DREAM-1K follows the released AutoDQ pipeline, using event extraction and bidirectional entailment to report precision, recall, and F1. Omni-Cloze uses the released cloze questions, answer options, and Reader protocol to measure visual, audio, audiovisual, and overall completion accuracy. For Daily-Omni, WorldSense, and Video-MME, a fixed GPT-5.6 caption-only Reader receives the frozen caption together with the original benchmark question and options, without access to the video or audio. Generation or parsing failures are counted as incorrect, and comparisons on shared videos use paired, video-level bootstrap estimates.

\subsection{Controlled Comparisons and Ablations}
\label{app:controlled-comparisons}
Data-construction comparisons change one stage at a time while holding the remaining training and evaluation conditions fixed. Video-filtering comparisons match the number, source, event category, and duration distribution of selected clips. Agent-supervision comparisons use the same videos and Captioner configuration, changing only the supervision captions. The Agent cascade comparison likewise evaluates direct captioning and \agent{} captions through the same frozen Reader, thereby measuring the system-level contribution of active evidence acquisition under their respective inference settings.

The \ppm{} ablation follows a paired design. Full-\ppm{} and no-\ppm{} captions are regenerated from the same video manifest with identical Planner, Finalizer, ALM/VLM access, decoding configuration, and output constraint; only the availability of the physical Observer changes. The resulting Captioners share initialization, optimization, preprocessing, and evaluation settings. Training and evaluation videos are separated by exact and perceptual duplicate checks before scoring.

\section{Limitations and Future Work}
\label{app:limitations-future}

Daily-Physics 50K and \bench{} focus on physical processes that are directly supported by audiovisual evidence. This scope enables traceable, scalable supervision over heterogeneous real-world videos without requiring force sensors, calibrated geometry, or simulator states, while emphasizing observable interactions, material responses, state changes, and physical outcomes. Future work can associate these evidence-rich captions with persistent object tracks, depth and 3D geometry, contact graphs, material estimates, and paired simulation or robot-interaction trajectories to obtain more structured physical world representations.

\agent{} uses modular evidence acquisition as an offline supervision engine and transfers the resulting knowledge to the tool-free \captionermodel{}, balancing detailed physical analysis with efficient deployment. Promising extensions include adaptive multi-frame observation, uncertainty-aware verification, reusable audiovisual representations, and evaluation of longer, egocentric, multi-object, action-conditioned, and counterfactual processes. These directions can strengthen world models and embodied agents by supporting interaction prediction, physically relevant experience retrieval, planning, failure diagnosis, and safety assessment in dynamic environments.

\FloatBarrier
\raggedbottom
\section{The Prompt Design of Agent}
\label{app:prompts}

The prompt settings of \agent{} are provided below. We present the complete configuration used when only the audio and visual Observers are available; the physical-perception Observer and its tool-specific instructions are intentionally excluded. Runtime limits are supplied through metadata and are not instantiated below.

\begin{plannerprompt}
\small
Plan Observer calls over the original media to collect audio, visual, textual, and temporal evidence for a detailed caption. Use the proxy clip only for navigation and ground the final caption in textual Observer evidence. Do not answer hidden questions or write the caption during planning.

\textbf{Tools.}
\begin{itemize}
\setlength{\itemsep}{1pt}\setlength{\parskip}{0pt}\setlength{\parsep}{0pt}
    \item \texttt{ALM}: audio only, covering speech, lyrics, music, sound effects, silence, applause, noise, speaker or audio transitions, and timing.
    \item \texttt{VLM}: visual only, covering people, objects, actions, states, colors, counts, positions, OCR, subtitles, interfaces, logos, transitions, camera changes, and the ending.
\end{itemize}

\textbf{Ask rules.}
\begin{itemize}
\setlength{\itemsep}{1pt}\setlength{\parskip}{0pt}\setlength{\parsep}{0pt}
    \item Obey the available-tool list and runtime values for \texttt{max\_ask\_rounds}, \texttt{max\_calls}, \texttt{min\_calls}, and \texttt{max\_questions\_per\_tool\_call}. Every call must specify a valid tool, start time, end time, and a nonempty list of neutral questions.
    \item Use event-centered windows and request timestamps for speech or lyrics, music changes, silence, and sound effects rather than only a transcript.
    \item Cover the opening, major action phases, transitions, later developments, and ending; use narrow follow-ups for exact actions, quoted phrases, cuts, and immediate aftermaths.
    \item Track the primary person or object, location, trajectory, tool use, and state change before secondary details. Group related neutral questions and avoid duplicate calls.
\end{itemize}

\textbf{Evidence priorities.}
\begin{enumerate}
\setlength{\itemsep}{1pt}\setlength{\parskip}{0pt}\setlength{\parsep}{0pt}
    \item Opening and ending states; exact event order; first, next, before, after, and immediately-after relations.
    \item Audio--visual alignment around speech, lyrics, sounds, music changes, actions, text appearances, and cuts.
    \item OCR, subtitles, interfaces, logos, names, numbers, charts, and scores, including when each appears, changes, persists, or disappears.
    \item Object count, color, position, contact, support, motion, transformation, outcome, and final location.
    \item High-risk evidence: commentary and scores, animation text and sound labels, narration after an introduction, and instructions aligned with hand or tool movement and its result.
\end{enumerate}

\textbf{QA-trap checklist.}
\begin{itemize}
\setlength{\itemsep}{1pt}\setlength{\parskip}{0pt}\setlength{\parsep}{0pt}
    \item Preserve specific and broader sound labels when both are supported.
    \item Treat backdrop graphics, lyric words, stylized letters, captions, and LED text as readable text.
    \item For ``during'' and ``immediately after,'' verify the overlapping visual and next state rather than a nearby event.
    \item Retain multiple plausible labels with calibrated uncertainty.
\end{itemize}

\textbf{Cross-modal rule.} Match key ALM evidence with concurrent VLM evidence and key VLM events with corresponding audio. Cover the complete timeline without becoming audio-only or visual-only.

\textbf{Inputs.} Proxy clip, \texttt{\{case\_context\}}, available-tool metadata, runtime limits including \texttt{max\_ask\_rounds}, \texttt{max\_calls}, \texttt{min\_calls}, and \texttt{max\_questions\_per\_tool\_call}, prior planning messages when present, and accumulated \texttt{\{evidence\_memory\}}.

\textbf{Output.} Return only a JSON object containing \texttt{tool\_calls}. Each call specifies \texttt{tool\_name}, \texttt{start\_time}, \texttt{end\_time}, and a list of \texttt{questions}. Return an empty list only when no further evidence is needed.

\textbf{JSON example.}\par
\texttt{\{}\par
\quad\texttt{"tool\_calls": [\{}\par
\qquad\texttt{"tool\_name": "ALM",}\par
\qquad\texttt{"start\_time": "00:00.000",}\par
\qquad\texttt{"end\_time": "00:10.000",}\par
\qquad\texttt{"questions": ["1. ..."]}\par
\quad\texttt{\}]}\par
\texttt{\}}
\end{plannerprompt}

\begin{firstprompt}
\small
\textbf{Round 1: build the evidence map.}

Watch the proxy video for navigation, then ask the Observers for original-quality evidence. Do not treat the proxy as the evidence source for the final caption.

\textbf{Planning requirements.}
\begin{enumerate}
\setlength{\itemsep}{1pt}\setlength{\parskip}{0pt}\setlength{\parsep}{0pt}
    \item Request a timestamped global audio map covering speech or lyrics by phrase, named entities, speaker changes, music, exact sound-effect labels, applause, silence, noise, and audio transitions. Add short ALM checks around salient moments instead of relying on the global map alone.
    \item Divide the visual timeline into the opening, major action phases, transitions, middle state, late events, and ending. Each VLM call should focus on a coherent event, shot group, text-card sequence, or action phase; exact actions and immediate-after states require localized evidence.
    \item Check OCR, subtitles, names, numbers, scores, labels, logos, interfaces, charts, products, and text changes. If a lyric, spoken phrase, or chant is prominent, verify whether matching or related words also appear on screen.
    \item Ask for the main people and objects, their initial positions, movement paths, entry or exit, interactions, visible outcomes, and states immediately before and after each major event.
    \item When an introduction precedes narration, identify when speech begins and what is visible at that moment. Distinguish live action from replay, demonstration, animation, and persistent overlays when relevant.
    \item For sports content, verify commentary, score or status, teams, clothing colors, the main action, and replay-versus-live order. For demonstrations or object manipulation, verify object identity, purpose, hand or tool movement, and the resulting state change.
    \item Align every salient sound or phrase with the concurrent visual state and the next visible change, and align every major visual action, text change, or cut with its accompanying audio.
\end{enumerate}

\textbf{Inputs.} Agent system prompt, proxy clip, \texttt{\{case\_context\}}, remaining-call metadata, available-tool metadata, and first-round planning metadata. No completed Observer evidence is available in this round.

\textbf{Output.} Return only the tool-call JSON defined by the system prompt. Do not stop after a coarse overview while meaningful intervals remain unchecked.

\textbf{JSON example.}\par
\texttt{\{}\par
\quad\texttt{"tool\_calls": [\{}\par
\qquad\texttt{"tool\_name": "VLM",}\par
\qquad\texttt{"start\_time": "00:10.000",}\par
\qquad\texttt{"end\_time": "00:18.000",}\par
\qquad\texttt{"questions": [}\par
\qquad\quad\texttt{"1. What changes in this interval?",}\par
\qquad\quad\texttt{"2. What is visible immediately afterward?"}\par
\qquad\texttt{]}\par
\quad\texttt{\}]}\par
\texttt{\}}
\end{firstprompt}

\begin{followupprompt}
\small
\textbf{Next evidence round.}

Inspect the accumulated Observer records and request only evidence that is missing, weakly localized, incomplete, or contradictory.

\textbf{Repair and completion rules.}
\begin{itemize}
\setlength{\itemsep}{1pt}\setlength{\parskip}{0pt}\setlength{\parsep}{0pt}
    \item Fill any uncovered opening, middle, late, or ending interval before finalization.
    \item If ALM reports an important phrase, sound, music change, applause, or silence, verify the overlapping visual and immediate aftermath. If VLM reports an important action, cut, text element, score change, or object-state change, inspect the corresponding audio.
    \item Refine generic audio descriptions into an exact type and order where possible. If evidence contains only a specific label, also check whether a broader answer-friendly category is supported; retain both when appropriate.
    \item Recheck titles, subtitles, lyric text, labels, interfaces, logos, charts, scores, numbers, products, and stylized backdrop text; determine whether each element newly appears, changes, disappears, or merely persists as an overlay.
    \item Repair transcript or commentary results that lack timestamps. Replace broad multi-event observations with the narrowest useful window around the relevant phrase, action, cut, or immediate-after state.
    \item Verify sequence-sensitive relations such as first, next, last, before, during, after, and immediately after rather than substituting a nearby event.
    \item Preserve the main foreground trajectory before secondary gestures or background overlays. Verify the purpose and visible result of tool use, object construction, medical or instructional actions, and other hand--object interactions.
    \item When Observers disagree about identity, team or player, score, purpose, object name, sound label, event order, or live-versus-replay status, request the narrowest targeted follow-up rather than carrying both claims forward without verification.
\end{itemize}

\textbf{Inputs.} Agent system prompt, proxy clip, \texttt{\{case\_context\}}, remaining-call metadata, \texttt{\{completed\_calls\}}, previous planning messages, and accumulated \texttt{\{evidence\_memory\}} containing the intervals, questions, Observer answers, and status metadata from earlier rounds.

\textbf{Output.} Return only the tool-call JSON. Return an empty \texttt{tool\_calls} list only when the timeline, cross-modal alignment, exact text, major object trajectories and states, contradictions, and ending are adequately supported.

\textbf{Continue example.}\par
\texttt{\{}\par
\quad\texttt{"tool\_calls": [\{}\par
\qquad\texttt{"tool\_name": "ALM",}\par
\qquad\texttt{"start\_time": "00:18.000",}\par
\qquad\texttt{"end\_time": "00:24.000",}\par
\qquad\texttt{"questions": [}\par
\qquad\quad\texttt{"1. What exact sound or speech occurs here, and in what order?"}\par
\qquad\texttt{]}\par
\quad\texttt{\}]}\par
\texttt{\}}

\textbf{Stop example.}\par
\texttt{\{}\par
\quad\texttt{"tool\_calls": []}\par
\texttt{\}}
\end{followupprompt}

\begin{audioprompt}
\small
\textbf{System message.} You are a precise audio Observer. Answer only from the provided audio segment.

\textbf{Rules.}
\begin{itemize}
\setlength{\itemsep}{1pt}\setlength{\parskip}{0pt}\setlength{\parsep}{0pt}
    \item Keep the same numbering as the questions.
    \item Be factual and conservative; explicitly correct unsupported assumptions in a question.
    \item Do not infer people, objects, actions, text, or any other visual content.
    \item Provide approximate timing and order inside the stated original-video interval.
    \item Report exact speech or lyrics when clear, named entities, speaker changes, music, exact sound-effect labels, applause, silence, noise, audio artifacts, and audio transitions.
    \item If a sound supports both a specific description and a broader category, preserve both without overstating certainty.
\end{itemize}

\textbf{User message.} Original video time range for this audio segment: \texttt{\{time\_range\}}. Answer from this audio only. Include approximate timing and order inside this range.

Questions:\par
\texttt{\{numbered\_questions\}}

\textbf{Inputs.} The cropped original-video audio segment, its original-video \texttt{\{time\_range\}}, the audio-only system message, and numbered \texttt{\{questions\}}.

\textbf{Output.} Return only final numbered answers using the same numbering as the questions. Do not return JSON, a caption, visual claims, planning text, or additional commentary.

\textbf{Example.}\quad\texttt{1. At approximately 00:03, ...}\par
\phantom{\textbf{Example.}\quad}\texttt{2. Immediately afterward, ...}
\end{audioprompt}

\clearpage
\begin{visualprompt}
\small
\textbf{System message.} You are a precise visual Observer. Answer only from the provided visual segment or sampled frames.

\textbf{Rules.}
\begin{itemize}
\setlength{\itemsep}{1pt}\setlength{\parskip}{0pt}\setlength{\parsep}{0pt}
    \item Keep the same numbering as the questions.
    \item Be factual and conservative; explicitly correct unsupported assumptions in a question.
    \item Do not infer speech, lyrics, music, sound effects, silence, or any other audio content.
    \item Provide approximate timing and order inside the stated original-video interval.
    \item Report people, objects, actions, states, counts, colors, positions, contact, entry or exit, movement trajectories, transformations, transitions, camera changes, and the ending state.
    \item Transcribe OCR, subtitles, labels, names, numbers, logos, scores, charts, and interface elements exactly when possible, and state whether text appears, changes, persists, or disappears.
    \item Separate the primary foreground event from secondary gestures, watermarks, and persistent background overlays.
    \item If temporal continuity or identity is unclear from the supplied segment or frames, state the uncertainty and prefer a weaker supported fact over a strong guess.
\end{itemize}

\textbf{User message.} Original video time range for this visual segment: \texttt{\{time\_range\}}. Answer from this visual input only. Include approximate timing and order inside this range.

Questions:\par
\texttt{\{numbered\_questions\}}

\textbf{Inputs.} The cropped original-video visual segment or sampled original-video frames, their original-video \texttt{\{time\_range\}}, the visual-only system message, frame timestamps when frames are used, and numbered \texttt{\{questions\}}.

\textbf{Output.} Return only final numbered answers using the same numbering as the questions. Do not return JSON, a caption, audio claims, planning text, or additional commentary.

\textbf{Example.}\quad\texttt{1. Near the start of the interval, ...}\par
\phantom{\textbf{Example.}\quad}\texttt{2. The visible text reads ``...'' and then disappears.}
\end{visualprompt}

\begin{finalprompt}
\small
You are the Finalizer. Given \texttt{\{case\_context\}} and the accumulated Observer records, write one evidence-rich caption and nothing else. Use the textual evidence; do not access the media or call tools at this stage.

\textbf{Requirements.}
\begin{enumerate}
\setlength{\itemsep}{1pt}\setlength{\parskip}{0pt}\setlength{\parsep}{0pt}
    \item Organize the caption chronologically from the initial scene and audio through meaningful events and transitions to the final state, using timestamps when supported.
    \item Convert useful, nonduplicate records into concrete prose, preserving actions, trajectories, object states and relations, counts, colors, positions, readable text, speech or lyrics, music, sounds, camera changes, and outcomes.
    \item For important phrases, sounds, text, actions, and cuts, describe the supported before, during, and after relations. Claim simultaneity only for overlapping timestamps; otherwise use ``before,'' ``after,'' or ``nearby.''
    \item Make the primary body, location, and object-state trajectory the backbone; add gestures, expressions, watermarks, and background overlays only after the main event is clear.
    \item Preserve exact OCR, text-like graphics, sound labels, and timing. Include both specific and broader sound categories when supported, and state how text appears, changes, persists, or disappears.
    \item Merge duplicates and order evidence by time. Prefer narrow, time-localized evidence over conflicting broad summaries, and do not restore discarded details from an isolated record.
    \item Resolve remaining conflicts with the safest supported wording and calibrated uncertainty. Express physical attributes qualitatively unless an exact value is printed or measured; never invent facts.
    \item End with a compact sentence covering the final visible state, continuing or stopped sound, and last action.
    \item Do not mention tools, prompts, Observer records, Evidence Memory, JSON, hidden questions, evaluation, or the proxy clip; do not repeat conclusions or pad the caption.
\end{enumerate}

\textbf{Inputs.} \texttt{\{case\_context\}}, accumulated audio and visual Observer records with their time ranges, questions, numbered answers, and status metadata, plus the available non-media planning context. Records marked unsuccessful or contradictory are not factual evidence unless confirmed elsewhere.

\textbf{Output.} Return only the final caption as chronological prose. Do not return JSON, headings, notes, explanations, or a list of tool records.
\end{finalprompt}

\clearpage
\clearpage

\clearpage

%% file: main.bbl
\begin{thebibliography}{47}
\providecommand{\natexlab}[1]{#1}
\providecommand{\url}[1]{\texttt{#1}}
\expandafter\ifx\csname urlstyle\endcsname\relax
  \providecommand{\doi}[1]{doi: #1}\else
  \providecommand{\doi}{doi: \begingroup \urlstyle{rm}\Url}\fi

\bibitem[Yang et~al.(2025)Yang, Wei, Li, Liu, Liu, Hu, He, Ju, Zhou, Liu,
  et~al.]{yang2025medaide}
Dingkang Yang, Jinjie Wei, Mingcheng Li, Jiyao Liu, Lihao Liu, Ming Hu, Junjun
  He, Yakun Ju, Wei Zhou, Yang Liu, et~al.
\newblock Medaide: Information fusion and anatomy of medical intents via
  {LLM}-based agent collaboration.
\newblock \emph{Information Fusion}, page 103743, 2025.

\bibitem[Yang et~al.(2026)Yang, Wei, Hu, Liu, Liu, Chen, Li, He, Zhou, Liu,
  et~al.]{yang2026toward}
Dingkang Yang, Jinjie Wei, Ming Hu, Jiyao Liu, Lihao Liu, Zhaoyu Chen,
  Mingcheng Li, Junjun He, Wei Zhou, Yang Liu, et~al.
\newblock Toward empathetic care: an {LLM}-based multi-intention recognition
  framework for mental health and complex medical queries.
\newblock \emph{IEEE Transactions on Affective Computing}, 2026.

\bibitem[Qian et~al.(2026)Qian, Chen, Liu, Sun, Yang, and
  Zhang]{qian2026spatialguard}
Ziyun Qian, Zizhi Chen, Yizhou Liu, Mingyang Sun, Dingkang Yang, and Lihua
  Zhang.
\newblock {SpatialGuard}: Harness-guided verifiable spatial reasoning for
  text-to-image generation.
\newblock \emph{arXiv preprint arXiv:2609.01582}, 2026.

\bibitem[Yang et~al.(2025)Yang, Xiao, Wei, Li, Chen, Li, and
  Zhang]{yang2025improving}
Dingkang Yang, Dongling Xiao, Jinjie Wei, Mingcheng Li, Zhaoyu Chen, Ke~Li, and
  Lihua Zhang.
\newblock Improving factuality in large language models via decoding-time
  hallucinatory and truthful comparators.
\newblock In \emph{Proceedings of the AAAI Conference on Artificial
  Intelligence}, volume~39, pages 25606--25614, 2025.

\bibitem[Han et~al.(2026)Han, Yang, Jiang, Liu, and Zhang]{han2026omnifysics}
Minghao Han, Dingkang Yang, Yue Jiang, Yizhou Liu, and Lihua Zhang.
\newblock {OmniFysics}: Towards physical intelligence evolution via omni-modal
  signal processing and network optimization.
\newblock \emph{arXiv preprint arXiv:2602.07064}, 2026.

\bibitem[Tung et~al.(2023)Tung, Ding, Chen, Bear, Gan, Tenenbaum, Yamins, Fan,
  and Smith]{tung2023physionpp}
Hsiao-Yu Tung, Mingyu Ding, Zhenfang Chen, Daniel Bear, Chuang Gan, Joshua~B.
  Tenenbaum, Daniel L.~K. Yamins, Judith~E. Fan, and Kevin~A. Smith.
\newblock {Physion++}: Evaluating physical scene understanding that requires
  online inference of different physical properties.
\newblock In \emph{Advances in Neural Information Processing Systems},
  volume~36, pages 67048--67068. Curran Associates, Inc., 2023.
\newblock Datasets and Benchmarks Track.

\bibitem[Zheng et~al.(2024)Zheng, Yan, Chen, Wang, Lim, Tenenbaum, and
  Gan]{zheng2024contphy}
Zhicheng Zheng, Xin Yan, Zhenfang Chen, Jingzhou Wang, Qin Zhi~Eddie Lim,
  Joshua~B. Tenenbaum, and Chuang Gan.
\newblock {ContPhy}: Continuum physical concept learning and reasoning from
  videos.
\newblock In \emph{Proceedings of the 41st International Conference on Machine
  Learning}, volume 235 of \emph{Proceedings of Machine Learning Research},
  pages 61526--61558. PMLR, 2024.

\bibitem[Gao et~al.(2024)Gao, Sarkar, Xia, Xiao, Wu, Ichter, Majumdar, and
  Sadigh]{gao2024physgrounded}
Jensen Gao, Bidipta Sarkar, Fei Xia, Ted Xiao, Jiajun Wu, Brian Ichter,
  Anirudha Majumdar, and Dorsa Sadigh.
\newblock Physically grounded vision-language models for robotic manipulation.
\newblock In \emph{2024 IEEE International Conference on Robotics and
  Automation}, pages 12462--12469. IEEE, 2024.

\bibitem[Chow et~al.(2025)Chow, Mao, Li, Seita, Guizilini, and
  Wang]{chow2025physbench}
Wei Chow, Jiageng Mao, Boyi Li, Daniel Seita, Vitor~Campagnolo Guizilini, and
  Yue Wang.
\newblock {PhysBench}: Benchmarking and enhancing {Vision-Language Models} for
  physical world understanding.
\newblock In \emph{The Thirteenth International Conference on Learning
  Representations}. OpenReview.net, 2025.

\bibitem[Krishna et~al.(2017)Krishna, Hata, Ren, Fei-Fei, and
  Niebles]{krishna2017dense}
Ranjay Krishna, Kenji Hata, Frederic Ren, Li~Fei-Fei, and Juan~Carlos Niebles.
\newblock Dense-captioning events in videos.
\newblock In \emph{Proceedings of the IEEE International Conference on Computer
  Vision}, pages 706--715, 2017.

\bibitem[Chen et~al.(2024)Chen, Wei, Li, Dong, Zhang, Zang, Chen, Duan, Lin,
  Tang, Yuan, Qiao, Lin, Zhao, and Wang]{chen2024sharegpt4video}
Lin Chen, Xilin Wei, Jinsong Li, Xiaoyi Dong, Pan Zhang, Yuhang Zang, Zehui
  Chen, Haodong Duan, Bin Lin, Zhenyu Tang, Li~Yuan, Yu~Qiao, Dahua Lin, Feng
  Zhao, and Jiaqi Wang.
\newblock {ShareGPT4Video}: Improving video understanding and generation with
  better captions.
\newblock arXiv preprint arXiv:2406.04325, 2024.

\bibitem[Chai et~al.(2025)Chai, Song, Du, Meng, Madhavan, Bar-Tal, Hwang, Xie,
  and Manning]{chai2025auroracap}
Wenhao Chai, Enxin Song, Yilun Du, Chenlin Meng, Vashisht Madhavan, Omer
  Bar-Tal, Jenq-Neng Hwang, Saining Xie, and Christopher~D. Manning.
\newblock {AuroraCap}: Efficient, performant video detailed captioning and a
  new benchmark.
\newblock In \emph{The Thirteenth International Conference on Learning
  Representations}, 2025.

\bibitem[Tang et~al.(2025)Tang, Li, Yang, Zhuang, Sun, Li, Ma, and
  Zhang]{tang2025videosalmonn2}
Changli Tang, Yixuan Li, Yudong Yang, Jimin Zhuang, Guangzhi Sun, Wei Li, Zejun
  Ma, and Chao Zhang.
\newblock {video-SALMONN 2}: Caption-enhanced audio-visual large language
  models.
\newblock arXiv preprint arXiv:2506.15220, 2025.

\bibitem[Chen et~al.(2025)Chen, Ding, Lin, Hua, Yao, Shi, Li, Zhang, Liu, Wan,
  Wang, and Tan]{chen2025avocado}
Xinlong Chen, Yue Ding, Weihong Lin, Jingyun Hua, Linli Yao, Yang Shi, Bozhou
  Li, Yuanxing Zhang, Qiang Liu, Pengfei Wan, Liang Wang, and Tieniu Tan.
\newblock {AVoCaDO}: An audiovisual video captioner driven by temporal
  orchestration.
\newblock arXiv preprint arXiv:2510.10395, 2025.

\bibitem[Ma et~al.(2026)Ma, Xu, Xing, Chu, Wang, He, Xu, Heng, Yu, Lin, Chng,
  and Chen]{ma2026omnicaptioner}
Ziyang Ma, Ruiyang Xu, Zhenghao Xing, Yunfei Chu, Yuxuan Wang, Jinzheng He, Jin
  Xu, Pheng-Ann Heng, Kai Yu, Junyang Lin, Eng~Siong Chng, and Xie Chen.
\newblock {Omni-Captioner}: Data pipeline, models, and benchmark for omni
  detailed perception.
\newblock In \emph{International Conference on Learning Representations}, 2026.

\bibitem[Tao et~al.(2025)Tao, Du, Yu, Wang, Liu, and Wang]{tao2025active}
Keda Tao, Wenjie Du, Bohan Yu, Weiqiang Wang, Jian Liu, and Huan Wang.
\newblock Active perception agent for omnimodal audio-video understanding.
\newblock arXiv preprint arXiv:2512.23646, 2025.

\bibitem[Cheng et~al.(2024)Cheng, Leng, Zhang, Xin, Li, Chen, Zhu, Zhang, Luo,
  Zhao, and Bing]{cheng2024videollama2}
Zesen Cheng, Sicong Leng, Hang Zhang, Yifei Xin, Xin Li, Guanzheng Chen,
  Yongxin Zhu, Wenqi Zhang, Ziyang Luo, Deli Zhao, and Lidong Bing.
\newblock {VideoLLaMA 2}: Advancing spatial-temporal modeling and audio
  understanding in video-{LLMs}.
\newblock arXiv preprint arXiv:2406.07476, 2024.

\bibitem[Xu et~al.(2025{\natexlab{a}})Xu, Guo, He, Hu, He, Bai, Chen, Wang,
  Fan, Dang, Zhang, Wang, Chu, and Lin]{xu2025qwen25omni}
Jin Xu, Zhifang Guo, Jinzheng He, Hangrui Hu, Ting He, Shuai Bai, Keqin Chen,
  Jialin Wang, Yang Fan, Kai Dang, Bin Zhang, Xiong Wang, Yunfei Chu, and
  Junyang Lin.
\newblock {Qwen2.5-Omni} technical report.
\newblock arXiv preprint arXiv:2503.20215, 2025{\natexlab{a}}.

\bibitem[Xu et~al.(2025{\natexlab{b}})Xu, Guo, Hu, Chu, Wang, He, Wang, Shi,
  He, Zhu, Lv, Wang, Guo, Wang, Ma, Zhang, Zhang, Hao, Guo, Yang, Zhang, Ma,
  Wei, Bai, Chen, Liu, Wang, Yang, Liu, Ren, Zheng, Men, Zhou, Yu, Yang, Yu,
  Zhou, and Lin]{qwen2025omni}
Jin Xu, Zhifang Guo, Hangrui Hu, Yunfei Chu, Xiong Wang, Jinzheng He, Yuxuan
  Wang, Xian Shi, Ting He, Xinfa Zhu, Yuanjun Lv, Yongqi Wang, Dake Guo,
  He~Wang, Linhan Ma, Pei Zhang, Xinyu Zhang, Hongkun Hao, Zishan Guo, Baosong
  Yang, Bin Zhang, Ziyang Ma, Xipin Wei, Shuai Bai, Keqin Chen, Xuejing Liu,
  Peng Wang, Mingkun Yang, Dayiheng Liu, Xingzhang Ren, Bo~Zheng, Rui Men, Fan
  Zhou, Bowen Yu, Jianxin Yang, Le~Yu, Jingren Zhou, and Junyang Lin.
\newblock {Qwen3-Omni} technical report.
\newblock arXiv preprint arXiv:2509.17765, 2025{\natexlab{b}}.

\bibitem[Yuan et~al.(2025)Yuan, Wang, Sun, Zhang, and Lin]{yuan2025tarsier2}
Liping Yuan, Jiawei Wang, Haomiao Sun, Yuchen Zhang, and Yuan Lin.
\newblock {Tarsier2}: Advancing large vision-language models from detailed
  video description to comprehensive video understanding.
\newblock arXiv preprint arXiv:2501.07888, 2025.

\bibitem[Wu et~al.(2025)Wu, Liu, Zhu, Zhou, and Shen]{wu2025ugc}
Peiran Wu, Yunze Liu, Zhengdong Zhu, Enmin Zhou, and Junxiao Shen.
\newblock {UGC-VideoCaptioner}: An omni ugc video detail caption model and new
  benchmarks.
\newblock arXiv preprint arXiv:2507.11336, 2025.

\bibitem[Wang et~al.(2026)Wang, Wang, Tang, Zhang, Cao, Bian, Zhang, Luo, Pan,
  Dong, Liu, and Zhang]{wang2026avscap}
Yanghai Wang, Jiahao Wang, Jiafu Tang, Yuanxing Zhang, Zhe Cao, Hanyan Bian,
  Zijie Zhang, Weiliang Luo, Zhiyu Pan, Zixuan Dong, Jiaheng Liu, and Zhaoxiang
  Zhang.
\newblock {AVSCap}: Orchestrating audio-visual synergy for omni-modal video
  captioning.
\newblock arXiv preprint arXiv:2607.12820, 2026.

\bibitem[Zhao et~al.(2026)Zhao, Ma, Huang, Fan, Li, Kang, Wei, Yang, and
  Tai]{zhao2026tca}
Chen Zhao, Jiajun Ma, Qilong Huang, Tiehan Fan, Hongyu Li, Zhuoliang Kang,
  Xiaoming Wei, Jian Yang, and Ying Tai.
\newblock Temporal and cross-modal alignment for enhanced audiovisual video
  captioning.
\newblock In \emph{European Conference on Computer Vision}, 2026.
\newblock To appear.

\bibitem[Zhou et~al.(2025)Zhou, Wang, Wu, and Jiang]{zhou2025dailyomni}
Ziwei Zhou, Rui Wang, Zuxuan Wu, and Yu-Gang Jiang.
\newblock {Daily-Omni}: Towards audio-visual reasoning with temporal alignment
  across modalities.
\newblock arXiv preprint arXiv:2505.17862, 2025.

\bibitem[Yu et~al.(2022)Yu, Wu, Liang, Salakhutdinov, and Morency]{yu2022pacs}
Samuel Yu, Peter Wu, Paul~Pu Liang, Ruslan Salakhutdinov, and Louis-Philippe
  Morency.
\newblock {PACS}: A dataset for physical audiovisual commonsense reasoning.
\newblock In \emph{Computer Vision -- ECCV 2022}, volume 13697 of \emph{Lecture
  Notes in Computer Science}, pages 292--309. Springer, 2022.

\bibitem[Krojer et~al.(2025)Krojer, Komeili, Ross, Garrido, Sinha, Ballas, and
  Assran]{krojer2025mvp}
Benno Krojer, Mojtaba Komeili, Candace Ross, Quentin Garrido, Koustuv Sinha,
  Nicolas Ballas, and Mahmoud Assran.
\newblock A shortcut-aware video-{QA} benchmark for physical understanding via
  minimal video pairs.
\newblock arXiv preprint arXiv:2506.09987, 2025.

\bibitem[Wu et~al.(2025)Wu, Li, Jin, Shi, KV, Raj, Sinha, Chen, Du, and
  Manocha]{wu2025mass}
Xiyang Wu, Zongxia Li, Jihui Jin, Guangyao Shi, Gouthaman KV, Vishnu Raj,
  Nilotpal Sinha, Jingxi Chen, Fan Du, and Dinesh Manocha.
\newblock {MASS}: Motion-aware spatial-temporal grounding for physics reasoning
  and comprehension in vision-language models.
\newblock arXiv preprint arXiv:2511.18373, 2025.

\bibitem[Cao et~al.(2024)Cao, Tang, Zhao, Guo, Liu, Zhang, Liu, Sun, Reid, and
  Liang]{cao2024physgame}
Meng Cao, Haoran Tang, Haoze Zhao, Hangyu Guo, Jiaheng Liu, Ge~Zhang, Ruyang
  Liu, Qiang Sun, Ian Reid, and Xiaodan Liang.
\newblock {PhysGame}: Uncovering physical commonsense violations in gameplay
  videos.
\newblock arXiv preprint arXiv:2412.01800, 2024.

\bibitem[Mak et~al.(2026)Mak, Zhu, Zhang, Li, Chi, Zhang, Wu, He, Fan, Lu, Ge,
  Fang, He, Lu, Xu, Zhang, Ni, Li, and Zhang]{mak2026physicsmind}
Chak-Wing Mak, Guanyu Zhu, Boyi Zhang, Hongji Li, Xiaowei Chi, Kevin Zhang,
  Yichen Wu, Yangfan He, Chun-Kai Fan, Wentao Lu, Kuangzhi Ge, Xinyu Fang,
  Hongyang He, Kuan Lu, Tianxiang Xu, Li~Zhang, Yongxin Ni, Youhua Li, and
  Shanghang Zhang.
\newblock {PhysicsMind}: Sim and real mechanics benchmarking for physical
  reasoning and prediction in foundational {VLMs} and world models.
\newblock arXiv preprint arXiv:2601.16007, 2026.

\bibitem[Li et~al.(2022)Li, Wei, Tian, Xu, Wen, and Hu]{li2022musicavqa}
Guangyao Li, Yake Wei, Yapeng Tian, Chenliang Xu, Ji-Rong Wen, and Di~Hu.
\newblock Learning to answer questions in dynamic audio-visual scenarios.
\newblock In \emph{Proceedings of the IEEE/CVF Conference on Computer Vision
  and Pattern Recognition}, pages 19086--19096, 2022.

\bibitem[Damen et~al.(2022)Damen, Doughty, Farinella, Furnari, Kazakos, Ma,
  Moltisanti, Munro, Perrett, Price, and Wray]{damen2022epickitchens100}
Dima Damen, Hazel Doughty, Giovanni~Maria Farinella, Antonino Furnari,
  Evangelos Kazakos, Jian Ma, Davide Moltisanti, Jonathan Munro, Toby Perrett,
  Will Price, and Michael Wray.
\newblock Rescaling egocentric vision: Collection, pipeline and challenges for
  {EPIC-KITCHENS-100}.
\newblock \emph{International Journal of Computer Vision}, 130\penalty0
  (1):\penalty0 33--55, 2022.

\bibitem[Zhang et~al.(2025)Zhang, Wu, Li, Li, Ma, Liu, and
  Li]{zhang2025llavavideo}
Yuanhan Zhang, Jinming Wu, Wei Li, Bo~Li, Zejun Ma, Ziwei Liu, and Chunyuan Li.
\newblock {LLaVA-Video}: Video instruction tuning with synthetic data.
\newblock \emph{Transactions on Machine Learning Research}, 2025.

\bibitem[Liu et~al.(2024)Liu, Li, Yu, Sheng, Wang, Li, and Yu]{liu2024luavs}
Chen Liu, Peike~Patrick Li, Qingtao Yu, Hongwei Sheng, Dadong Wang, Lincheng
  Li, and Xin Yu.
\newblock Benchmarking audio visual segmentation for long-untrimmed videos.
\newblock In \emph{Proceedings of the IEEE/CVF Conference on Computer Vision
  and Pattern Recognition}, pages 22712--22722, 2024.

\bibitem[Yang et~al.(2025)Yang, Li, Chen, Chen, Yao, and
  Lin]{yang2025videomind}
Baoyao Yang, Wanyun Li, Dixin Chen, Junxiang Chen, Wenbin Yao, and Haifeng Lin.
\newblock {VideoMind}: An omni-modal video dataset with intent grounding for
  deep-cognitive video understanding.
\newblock arXiv preprint arXiv:2507.18552, 2025.

\bibitem[Wang et~al.(2025)Wang, Ma, Cao, Zheng, Zhang, Feng, Liu, Ma, Cheng,
  Leng, Yin, and Liang]{wang2025wisa}
Jing Wang, Ao~Ma, Ke~Cao, Jun Zheng, Zhanjie Zhang, Jiasong Feng, Shanyuan Liu,
  Yuhang Ma, Bo~Cheng, Dawei Leng, Yuhui Yin, and Xiaodan Liang.
\newblock {WISA}: World simulator assistant for physics-aware text-to-video
  generation.
\newblock arXiv preprint arXiv:2503.08153, 2025.

\bibitem[{Hugging Face}(2024)]{huggingface2024finevideo}
{Hugging Face}.
\newblock {FineVideo}.
\newblock Hugging Face dataset card, 2024.

\bibitem[Geng et~al.(2023)Geng, Wang, Duan, Cong, and Zheng]{geng2023unav100}
Tiantian Geng, Teng Wang, Jinming Duan, Runmin Cong, and Feng Zheng.
\newblock Dense-localizing audio-visual events in untrimmed videos: A
  large-scale benchmark and baseline.
\newblock In \emph{Proceedings of the IEEE/CVF Conference on Computer Vision
  and Pattern Recognition}, pages 22942--22951, 2023.

\bibitem[Li et~al.(2026)Li, Zhang, Long, Chen, Song, Bai, Yang, Xie, Yang, Liu,
  Zhou, and Lin]{li2026qwen3vlembedding}
Mingxin Li, Yanzhao Zhang, Dingkun Long, Keqin Chen, Sibo Song, Shuai Bai,
  Zhibo Yang, Pengjun Xie, An~Yang, Dayiheng Liu, Jingren Zhou, and Junyang
  Lin.
\newblock {Qwen3-VL-Embedding} and {Qwen3-VL-Reranker}: A unified framework for
  state-of-the-art multimodal retrieval and ranking.
\newblock arXiv preprint arXiv:2601.04720, 2026.

\bibitem[Yao et~al.(2023)Yao, Zhao, Yu, Du, Shafran, Narasimhan, and
  Cao]{yao2023react}
Shunyu Yao, Jeffrey Zhao, Dian Yu, Nan Du, Izhak Shafran, Karthik~R.
  Narasimhan, and Yuan Cao.
\newblock {ReAct}: Synergizing reasoning and acting in language models.
\newblock In \emph{The Eleventh International Conference on Learning
  Representations}. OpenReview.net, 2023.

\bibitem[Wang et~al.(2024)Wang, Yuan, Zhang, and Sun]{wang2024tarsier}
Jiawei Wang, Liping Yuan, Yuchen Zhang, and Haomiao Sun.
\newblock {Tarsier}: Recipes for training and evaluating large video
  description models.
\newblock arXiv preprint arXiv:2407.00634, 2024.

\bibitem[{OpenAI}(2024)]{openai2024gpt4o}
{OpenAI}.
\newblock {GPT-4o} system card.
\newblock arXiv preprint arXiv:2410.21276, 2024.

\bibitem[{Google}(2026{\natexlab{a}})]{google2026gemini35flash}
{Google}.
\newblock {Gemini 3.5 Flash}.
\newblock Google AI for Developers model documentation, 2026{\natexlab{a}}.

\bibitem[{Google}(2026{\natexlab{b}})]{google2026gemini31pro}
{Google}.
\newblock {Gemini 3.1 Pro} preview.
\newblock Google AI for Developers model documentation, 2026{\natexlab{b}}.

\bibitem[Cui et~al.(2026)]{cui2026minicpmo45}
Junbo Cui et~al.
\newblock {MiniCPM-o 4.5}: Towards real-time full-duplex omni-modal
  interaction.
\newblock arXiv preprint arXiv:2604.27393, 2026.

\bibitem[Hong et~al.(2026)Hong, Yan, Cai, Jiang, Hu, and
  Xie]{hong2026worldsense}
Jack Hong, Shilin Yan, Jiayin Cai, Xiaolong Jiang, Yao Hu, and Weidi Xie.
\newblock {WorldSense}: Evaluating real-world omnimodal understanding for
  multimodal {LLMs}.
\newblock In \emph{International Conference on Learning Representations}, 2026.

\bibitem[Fu et~al.(2025)Fu, Dai, Luo, Li, Ren, Zhang, Wang, Zhou, Shen, Zhang,
  Chen, Li, Lin, Zhao, Li, Xu, Zheng, Chen, Shan, He, and Sun]{fu2025videomme}
Chaoyou Fu, Yuhan Dai, Yongdong Luo, Lei Li, Shuhuai Ren, Renrui Zhang, Zihan
  Wang, Chenyu Zhou, Yunhang Shen, Mengdan Zhang, Peixian Chen, Yanwei Li,
  Shaohui Lin, Sirui Zhao, Ke~Li, Tong Xu, Xiawu Zheng, Enhong Chen, Caifeng
  Shan, Ran He, and Xing Sun.
\newblock {Video-MME}: The first-ever comprehensive evaluation benchmark of
  multi-modal {LLMs} in video analysis.
\newblock In \emph{Proceedings of the IEEE/CVF Conference on Computer Vision
  and Pattern Recognition}, pages 24108--24118, 2025.

\bibitem[Chen et~al.(2024)Chen, Siarohin, Menapace, Deyneka, Chao, Jeon, Fang,
  Lee, Ren, Yang, and Tulyakov]{chen2024panda70m}
Tsai-Shien Chen, Aliaksandr Siarohin, Willi Menapace, Ekaterina Deyneka,
  Hsiang-Wei Chao, Byung~Eun Jeon, Yuwei Fang, Hsin-Ying Lee, Jian Ren,
  Ming-Hsuan Yang, and Sergey Tulyakov.
\newblock {Panda-70M}: Captioning 70m videos with multiple cross-modality
  teachers.
\newblock arXiv preprint arXiv:2402.19479, 2024.

\end{thebibliography}
